\documentclass{article}

\usepackage{amsmath,amsfonts,bm}

\def\eqref#1{equation~\ref{#1}}

\def\ceil#1{\lceil #1 \rceil}

\def\1{\bm{1}}

\def\mB{{\bm{B}}}

\def\mL{{\bm{L}}}

\DeclareMathAlphabet{\mathsfit}{\encodingdefault}{\sfdefault}{m}{sl}
\SetMathAlphabet{\mathsfit}{bold}{\encodingdefault}{\sfdefault}{bx}{n}

\newcommand{\R}{\mathbb{R}}

\newcommand{\Var}{\mathrm{Var}}
\newcommand{\diag}{\mathrm{diag}}

\DeclareMathOperator*{\argmin}{arg\,min}

\DeclareMathOperator{\vect}{vec}
\DeclareMathOperator{\dist}{dist}

\usepackage{microtype}
\usepackage{graphicx}
\usepackage{subfigure}
\usepackage{booktabs} 
\usepackage{amsmath,amssymb,amsthm}
\usepackage{caption}
\usepackage{physics}
\usepackage{algorithm}
\usepackage{algorithmic}
\usepackage{physics}
\usepackage{natbib}
\usepackage[title]{appendix}
\usepackage{adjustbox}
\usepackage{amsthm,amsmath,amssymb}
\usepackage{xspace}
\usepackage{tabularx}
\usepackage{soul}

\newcommand{\dpclip}{\textsc{Dp-CLIP}\xspace}
\newcommand{\dpblip}{\textsc{Dp-BLIP}\xspace}
\newcommand{\bc}{\color{blue}}

\def\mL{\mathcal{L}_\text{L}}
\def\mB{\mathcal{B}}
\def\pG{\partial_G}

\newcommand{\indep}{\perp \!\!\! \perp}

\usepackage{hyperref}

\usepackage[accepted]{icml2024}

\usepackage{amsmath}
\usepackage{amssymb}
\usepackage{mathtools}
\usepackage{amsthm}

\usepackage[capitalize,noabbrev]{cleveref}

\theoremstyle{plain}
\newtheorem{theorem}{Theorem}[section]
\newtheorem{proposition}[theorem]{Proposition}
\newtheorem{lemma}[theorem]{Lemma}
\newtheorem{corollary}[theorem]{Corollary}
\theoremstyle{definition}
\newtheorem{definition}[theorem]{Definition}
\newtheorem{assumption}[theorem]{Assumption}
\theoremstyle{remark}
\newtheorem{remark}[theorem]{Remark}

\newcommand{\wz}[1]{\todo{[Wanrong: #1]}}

\usepackage[textsize=tiny]{todonotes}

\icmltitlerunning{Safeguarding Data in Multimodal AI}

\begin{document}

\twocolumn[
\icmltitle{Safeguarding Data in Multimodal AI:\\A Differentially Private Approach to CLIP Training}



\icmlsetsymbol{equal}{*}

\begin{icmlauthorlist}
\icmlauthor{Alyssa Huang}{equal,yyy}
\icmlauthor{Peihan Liu}{equal,yyy}
\icmlauthor{Ryumei Nakada}{equal,xxx}
\icmlauthor{Linjun Zhang}{xxx}
\icmlauthor{Wanrong Zhang}{yyy}
\end{icmlauthorlist}

\icmlaffiliation{yyy}{Harvard University}
\icmlaffiliation{xxx}{Rutgers Univeristy}

\icmlcorrespondingauthor{Wanrong Zhang}{imwanrongz@gmail.com}
\icmlcorrespondingauthor{Linjun Zhang}{linjun.zhang@rutgers.edu}

\icmlkeywords{Machine Learning, ICML}

\vskip 0.3in
]



\printAffiliationsAndNotice{\icmlEqualContribution} 

\begin{abstract}
The surge in multimodal AI's success has sparked concerns over data privacy in vision-and-language tasks. While CLIP has revolutionized multimodal learning through joint training on images and text, its potential to unintentionally disclose sensitive information necessitates the integration of privacy-preserving mechanisms. We introduce a differentially private adaptation of the Contrastive Language-Image Pretraining (CLIP) model that effectively addresses privacy concerns while retaining accuracy. 
Our proposed method, \dpclip, is rigorously evaluated on benchmark datasets encompassing diverse vision-and-language tasks such as image classification and image captioning. 
We demonstrate that our approach retains performance on par with the standard non-private CLIP model. Furthermore, we analyze our proposed algorithm under linear representation settings. We derive the convergence rate of our algorithm and show a trade-off between utility and privacy when gradients are clipped per-\textit{batch} and the loss function does not satisfy smoothness conditions assumed in the literature for the analysis of DP-SGD.
\end{abstract}

\section{Introduction}
\label{sec:intro}

The field of vision-language tasks has witnessed a revolutionary breakthrough with the introduction of Contrastive Language–Image Pre-training (CLIP) \citep{radford2021learning} by OpenAI. It has redefined the benchmarks of various downstream vision-language tasks through its exceptional flexibility and remarkable zero-shot learning ability \citep{liang2021llm, Dorbala2022clipnav}. CLIP and its successors have been widely used in vision-language tasks such as semantic segmentation, image generation from captions, video summarization, and visual question answering \citep{generating2021,narasimhan2021clip,yao2021filip,li2021supervision,xu2022multimodal,wang2022clip}.

However, representations from contrastive learning are rich in information, which might be able to unintentionally memorize more personal sensitive information.
\citet{carlini2023extracting} shows that diffusion models, which are built upon CLIP, can memorize individual images in the training data and emit them at generation time, and is even less private compared to prior generative models such as GANs.
\citet{encodermi} successfully conducts membership inference attacks on the image encoder trained from CLIP. Furthermore, \citet{attributeinference} demonstrates that SimCLR, another framework that employs contrastive learning for image representations, is more susceptible to attribute inference attacks compared to standard supervised models. 

Mitigating these privacy vulnerabilities requires privacy-preserving training methods for multimodal models. The field of {\em differential privacy} (DP) has emerged as the gold standard for privacy-preserving data analytics. It is a mathematical notion of privacy, which ensures that the output distribution of computation is robust to one person's data. DP is typically achieved by introducing randomness in the computation and DP techniques have been employed in the context of unimodal models  \citep{xu2022multimodalsurvey,Basu2021BenchmarkingDP,Hoory2021LearningAE, yu2022differentially, li2022large}. However, differentially private training of multimodal models is currently understudied. \citep{carlini2023extracting} shows directly applying DP-SGD \citep{abadi2016deep} to diffusion models on CIFAR-10 causes the training consistently diverge, even when low privacy guarantees.

In this paper, we propose \dpclip, which safeguards vision-language tasks by learning differentially private image and text representations. These representations can then be safely used for various downstream multimodal tasks. It is worth emphasizing that our work represents one of the first attempts to incorporate privacy into multimodal models, thus paving the way for enhanced privacy protection in vision-language tasks. We hope that, by leveraging the information contained in the pretrained embeddings using public data, our fine-tuned private representations can maintain their utility in performing various downstream tasks. 
Since the CLIP loss function 
involves contrasting data from different modalities, the standard DP deep learning approach, DP-SGD based on \textit{per-sample} clipping, cannot be directly applied.  To address this challenge, we employ \textit{per-batch} clipping and show that our algorithm achieves high accuracy while protecting the desired level of privacy, from both theoretical and empirical perspectives.

To demonstrate the strength of our proposed method, we evaluate our differentially private CLIP, \dpclip, on benchmark datasets encompassing diverse vision-and-language tasks such as image classification and image captioning. 
The findings demonstrate that our privacy-aware approach retains performance on par with the standard CLIP model while significantly reducing the risk of data exposure. 

Furthermore, we theoretically derive the privacy-utility trade-off of \dpclip under linear representation settings.
Previous works analyzed the convergence rate of DP-SGD under certain smoothness conditions \citep{yu2019gradient,bassily2019private,feldman2020private,chen2020understanding,yang2022normalized,bu2022automatic,fangimproved}. However, we note that the CLIP loss function is not convex nor satisfy smoothness conditions in these literature.
We also note that we deal with \textit{per-batch} clipping instead of \textit{per-sample} clipping analyzed in \citet{abadi2016deep,chen2020understanding,yang2022normalized,bu2022automatic,fangimproved}. Although the loss function does not globally behave well, we exploit the fact that the linearized loss is locally smooth and strongly convex, and provide a probabilistic bound with linear convergence in the number of iterations.

The rest of this paper is organized as follows. We provide the necessary background for interpreting our results in Section \ref{sec:prelim}. In Section \ref{sec:dpclip}, we introduce our algorithm \dpclip. We present our experiments setup and empirical results in Section \ref{sec:exp}, followed by our theoretical analysis in Section \ref{sec: theory}. Finally, we conclude with our discussion in Section \ref{sec: discussion}.

\section{Preliminaries}\label{sec:prelim}

\noindent\textbf{Differential Privacy}
{\em Differential Privacy} is a mathematical definition of privacy. 
It is used to enable the analysis of sensitive data while preserving the privacy of individuals. In essence, differential privacy requires a mechanism's outputs on two adjacent datasets, which differ in one arbitrary person's data, to be indistinguishable. This is achieved by introducing randomness into the computation.
\begin{definition}[Differential Privacy \citep{dwork2006calibrating}]
A randomized algorithm $M:\mathcal{X}\rightarrow\mathcal{Y}$ preserves $(\epsilon,\delta)$-differential privacy, if for all adjacent $X,X'\in\mathcal{X}$ such that $\forall S \subseteq \mathcal{Y}$, we always have
\begin{equation}
    Pr[M(X)\in S]\leq e^\epsilon Pr[M(X')\in S] + \delta.
\end{equation}
\end{definition}

Differentially private mechanisms typically add noise that scales with the {\em sensitivity} of the function being evaluated. The sensitivity of a function $f$ is defined as the maximum change in $f$ between two neighboring (adjacent) sets: $\Delta f = \max_{\text{adjacent }X,X' } | f(X) - f(X')|$.
The {\em Gaussian mechanism} with parameters $(\epsilon, \delta)$ takes in a function $q$, dataset $X$, and outputs $f(X)+\mathcal{N}(0,\sigma^2)$, where $\sigma=\sqrt{2\log(1.25/\delta)}\Delta f/\epsilon$. This canonical mechanism serves as a base for DP-SGD, which is the current state-of-the-art framework for DP deep learning.

Existing works have explored the effects of clipping and noise addition on the convergence and performance of DP deep learning models. \citet{bu2021convergence} analyzed the impact of \textit{per-sample} clipping and noise addition on the convergence of DP deep learning models, characterizing these effects through training dynamics under the neural tangent kernel setting. They also introduced a new technique called \textit{global clipping} that improves the convergence rate of DP-SGD. 
\citet{wang2019differentially} studied the convergence properties of DP-SGD and showed that it converges to an approximate local minimum with high probability. \citet{imtiaz2017differentially} conducted differentially private canonical correlation analysis (DP-CCA) experiments  to evaluate the effectiveness of differential privacy in preventing membership inference attacks.
\citet{yu2019gradient} showed that gradient perturbation is efficient and accurate when the sample size is large and suggested that gradient perturbation may be combined with other differential privacy techniques to achieve even better results.

\noindent\textbf{CLIP} The \begin{em}Contrastive Language-Image Pre-training\end{em} (CLIP) \citet{radford2021learning} is a multimodal vision and language model, trained on image-text pairs. It is used to produce embeddings for both texts and images. Specifically, let $f: \R^{d_1} \to \R^r$ and $\tilde f: \R^{d_2} \to \R^r$ be the dual encoders.
Given pairs of data $\{(x_i, \tilde x_i)\}_{i \in [n]} \subset \R^{d_1+d_2}$, the CLIP loss can be formulated as follows:
\begin{align}
    \mathcal{L}(f, \tilde f) = &- \frac{1}{n} \sum_{i \in [n]} \log \frac{e^{s_{ii}/\tau}}{\sum_{j \in [n]} e^{s_{ij}/\tau}} \nonumber\\
    &- \frac{1}{n} \sum_{i \in [n]} \log \frac{e^{s_{ii}/\tau}}{\sum_{j \in [n]} e^{s_{ji}/\tau}},\label{loss: CLIP}
\end{align}
where $s_{ij} := \text{Sim}(f(x_i), \tilde f(\tilde x_j))$ is the cosine similarity of $x_i$ and $\tilde x_j$ measured in the feature space, and $\tau > 0$ is the temperature parameter.
The loss in \eqref{loss: CLIP} is a type of {\em contrastive loss}, that trains encoders by classifying based on whether a pair is observed or artifically paired.
Intuitively, this loss maps images and texts, that refer to the same object, to vectors with high cosine similarity and map images and texts that are unrelated to vectors with low cosine similarity.
We provide the details of the pretraining process to the Appendix \ref{app.prelim}.

\section{\dpclip: Private and Accurate Representations}
\label{sec:dpclip}
In this section, we introduce our \dpclip that incorporates privacy into the CLIP model, presented formally in Algorithm \ref{alg:DP-GD}. Given initial representation functions, the algorithm trains encoders based on per-batch noisy SGD to ensure privacy. 
The CLIP loss \citep{radford2021learning} is the contrastive cross-entropy loss function computed from pseudo-labels distinguishing whether a pair is observed or artificially generated.
Since the CLIP loss function, by definition, contains the similarity between data from multiple pairs, it cannot be written as a sum of losses of individual pairs. Thus we cannot directly apply per-sample clipping technique as in the original DP-SGD. Instead, we employ \textit{per-batch} clipping, and we show that can still guarantee the desired level of privacy, which we discuss in details below.

More specifically, we consider the following setups. Let $f_{\theta_1}$ and $\tilde f_{\theta_2}$ be the dual encoders to be trained, parameterized by $\theta_1$ and $\theta_2$, respectively.
Let $\theta^{(t)} = (\theta_1^{(t)}, \theta_2^{(t)})$ be the parameters at $t$-th iteration. We obtain a sequence of parameters $(\theta^{(t)})_{t=1}^T$ through mini-batch stochastic gradient descent, where $T$ is the number of iterations.
For any subset $\mB \subset [n]$, define $\mathcal{L}(\ \cdot\ ; \mB)$ be the loss \ref{loss: CLIP} computed only with pairs $\{(x_i, \tilde x_i)\}_{i\in\mB}$.
At iteration $t$, we uniformly sub-sample a mini-batch $\mB^{(t)}$ of size $b$ from $[n]$ and compute the partial derivative of $\mathcal{L}(f_{\theta_1}, \tilde f_{\theta_2}; \mB^{(t)})$, the loss computed with mini-batch, with respect to $\theta$ evaluated at $\theta_1 = \theta_1^{(t)}$ and $\theta_2 = \theta_2^{(t)}$. We denote this as $\partial_{\theta}\mathcal{L}(f_{\theta_1^{(t)}}, \tilde f_{\theta_2^{(t)}}; \mB)$ for brevity.
Then, we clip the mini-batch gradient with a clipping threshold $c > 0$. Let $h^{(t)} = \min\{1,\ c/\|\partial_{\theta}\mathcal{L}(f_{\theta_1^{(t)}}, \tilde f_{\theta_2^{(t)}}; \mB^{(t)})\|_F\}$, and we update $\theta^{(t)}$ as follows:
\begin{align*}
    \theta^{(t+1)} &= \theta^{(t)} - \eta \qty{h^{(t)} \partial_{\theta}\mathcal{L}\bigl(f_{\theta_1^{(t)}}, \tilde f_{\theta_2^{(t)}}; \mB^{(t)}\bigr) + \sigma c \Gamma^{(t)} },
\end{align*}
where $\eta > 0$ is the learning rate and $\vect(\Gamma^{(t)}) \sim N(0, I_{r d})$ is the noise added to ensure differential privacy. We formally present the algorithm in \ref{alg:DP-GD}.

\begin{algorithm}
\caption{\dpclip}\label{alg:DP-GD}
\begin{algorithmic}[1]
\STATE \textbf{Input:} observed pairs of data $\{(x_i, \tilde x_i)\}_{i=1}^n$, number of iterations $T$, noise scale $\sigma$, clipping threshold $c$, learning rate $\eta$, mini-batch size $b$,
initial parameters $\theta^{(0)} = (\theta_1^{(0)}, \theta_2^{(0)})$. 
\FOR{$t \in \{0, \dots, T-1\}$}
\STATE \textsf{Sample mini-batch:}
\\ \qquad Uniformly sample $\mB^{(t)}$ of size $b$ from $[n]$.
\STATE \textsf{Compute mini-batch gradient:}
\\ \qquad $g^{(t)} \leftarrow \partial_{\theta}\mathcal{L}(f_{\theta_1^{(t)}}, \tilde f_{\theta_2^{(t)}}; \mB)$. 
\STATE \textsf{Clip Gradient:}
\\ \qquad $\bar g^{(t)} \leftarrow \min\{1, \ c/\|g^{(t)}\|_F\} g^{(t)}$.
\STATE \textsf{Add Noise:}
\\ \qquad $\tilde g^{(t)} \leftarrow \bar g^{(t)} + \sigma c \mathcal{N}(0, I)$.
\STATE \textsf{Descent:}
\\ \qquad $\theta^{(t+1)} \leftarrow \theta^{(t)} - \eta \tilde g^{(t)}$.
\ENDFOR
\STATE \textbf{Return} $\theta^{(T)}$.
\end{algorithmic}
\end{algorithm}

We show that this algorithm is differentially private in a \textit{per-sample} level. 
In particular, the following result suggests that by carefully choosing $\sigma$, it can achieve desired differential privacy guarantee.

\begin{proposition}\label{prop: dp}
    Choose $b < n/10$.
    There exists universal constants $C_\epsilon, C_\sigma > 0$ such that
    for any $\epsilon \leq C_\epsilon b^2 T/n^2$ and $\delta > 0$, \dpclip is $(\epsilon,\delta)$-differentially private if we choose $\sigma \geq C_\sigma \sqrt{T \log(1/\delta)} / (n \epsilon)$.
\end{proposition}
Proposition \ref{prop: dp} states that Gaussian mechanism applied on mini-batch level with calibrated noise satisfies differential privacy in a \textit{per-sample} level.
The proof is deferred to the Appendix, where the technique follows directly from existing results \citep{bun2018composable,yang2022normalized}. Although the privacy analysis is straightforward, our main goal is to demonstrate the efficiency of this simple approach for CLIP training, which we will show by both our experiments and theoretical analysis. We note that the theoretical analysis of utility is quite challenging because the technical tools employed in the existing literature to study DP-SGD cannot be directly applied to the \textit{non-decomposable} loss, which requires us to develop novel technical tools. We defer this part of theory to Section \ref{sec: theory}.

\begin{remark}
    For implementation of \dpclip, we can utilize the existing microbatch SGD \citep{mcmahan2018general} implementation in Tensorflow privacy library by setting the number of microbatches to $1$.
    However, we note that the rationale behind our method and microbatch SGD differs.
    The microbatch clipping approach is designed to consider the setting of multiple queries per user and enhance training efficiency by leveraging larger number of microbatches.
    In fact, setting the number of microbatches to $1$ in microbatch SGD is atypical \citep{mcmahan2018general,bu2020deep,dupuy2022efficient}.
    On the other hand, our method specifically implements per-batch clipping due to the fact that the CLIP loss is non-decomposable.
\end{remark}

\section{Experiments}
\label{sec:exp}
In this section, we evaluate our \dpclip\ on image classification and image captioning tasks, and we defer the Visual Question Answering (VQA) task to Appendix \ref{sec:VQA}. We first introduce the training details in Section \ref{sec.setup}, then we provide detailed experimental results in Section \ref{sec.classification} and Section \ref{sec:imgcap}. Our code is available in the supplementary materials. 

\subsection{Experiments Setup}\label{sec.setup}

We acknowledge that in all of our tasks, we use a pretrained model as an initialization to reduce computational costs, and we continue to privately train them using contrastive loss on private datasets (e.g., MNIST, which is treated as private in our paper). We note that our proposed algorithm and the associated theory are designed for \dpclip training from scratch.



For image classification, we apply Algorithm \ref{alg:DP-GD} to image and caption pairs with prompt engineering. We then calculate the cosine similarity between the image and text encoders for classes on the testing set. The predicted class is the one with the highest cosine similarity, following \citet{radford2021learning}.
We also demonstrate that our \dpclip framework can be applied to a broad class of Vision-Language Pre-training models and can therefore be leveraged to perform more complex vision-language tasks. One such task is image captioning. To this end, for image captioning, we use the pretrained {\em Bootstrapping Language Image Pre-training} (BLIP) \citet{li2022blip}, which is an extension of CLIP. We provide the background of BLIP in Appendix \ref{app:exp}. We apply the same per-batch clipping framework and noise injection process described in Algorithm \ref{alg:DP-GD} on BLIP loss function. 
We call this \dpblip. 



\paragraph{Datasets}
For image classification, we first consider four benchmark image classification datasets , namely MNIST \citep{mnistdataset}, Fashion-MNIST \citep{fashionmnistdataset}, CIFAR-10 \citep{cifar10dataset}, and SVHN \citep{svhndataset}. To represent a more realistic application setting, we consider private datasets that have significant distribution shit with respect to the pre-training set \cite{kramerkamathcarlini2022}. The German Traffic Sign Recognition Benchmark (GTSRB) dataset \cite{gtsrbdataset} is a dataset with 43 traffic sign categories. GTSRB does not have too much overlap with ImageNet-1K since Image-1K lumps all of its 43 traffic sign categories into a single label \cite{pinto2023pillar}. In addition, we also consider CIFAR-100 \cite{cifar100dataset}, a dataset where there is relatively lower number of training samples per class and DP is known to be harder. 
For different datasets, we use different prompts to feed into the model, and we will discuss the details below.


For image captioning task, we employ partial data from the vizwiz image captioning dataset \citep{vizwiz}, which consists of a training set of 5k image-caption pairs (out of 23k) and a test set of 1k image-caption pairs (out of 8k). We use vizwiz evaluation api to evaluate our results, including BLEU, Rouge$\_$L, CIDEr, and SPICE.


\paragraph{Model Architecture}
For image classification, the base model is OpenAI's CLIP model pretrained on the ImageNet dataset \citep{deng2009imagenet}. It consist of a \textit{ViT-L/14-336px} Transformer architecture as an image encoder and a masked self-attention Transformer as a text encoder. The vision Transformer consists of 24 layers with width of 1024 and 16 heads. The text Transformer consists of 12 layers with width of 768 and 12 heads.

For image captioning, the base model is BLIP pretrained on two human-annotated datasets, COCO \citep{coco} and Visual Genome \citep{VG}, and
three web datasets, Conceptual Captions \citep{CCdata}, Conceptual 12M \citep{CC12}, and SBU captions \citep{SBU}, consisting of an unimodal image encoder, an image-grouded text encoder and an answer decoder, as illustrated in Figure \ref{fig:clip-pretraining-blip_vqa} in Appendix \ref{app:exp}.

\paragraph{Training Details}
The value of the noise multiplier, $\sigma$, is determined by the size of the training set $n$, the batch size $b$, the number of iterations $T$, and privacy parameters $(\epsilon, \delta)$. Proposition \ref{prop: dp} provides a guidance on the choice of these parameters. We use TensorFlow Privacy\footnote{\url{https://github.com/tensorflow/privacy}} for privacy accounting (RDP accountant.) Throughout the experiments, we set $\delta = 1/(2n)$ where $n$ is the size of the training set. All our models are implemented in PyTorch \citep{paszke2019pytorch} using one NVIDIA A100 80G GPU. 

\subsection{Image Classification Results}\label{sec.classification}
To obtain optimal results, we conduct hyperparameter tuning and prompt engineering. Because the baseline methods in \citet{temperedsigmoid} and \citet{dpkip} did not account for the privacy loss on tuning, we do the same to ensure a fair comparison. We set the learning rate $\eta=10^{-5}$, since we find it yields the best performance across the datasets. 
We vary the clipping threshold $c$ from $0.1$ to $1$, and batch size $b$ from 16 to 128. We train the model for 15 to 30 epochs. Regarding prompt engineering, we employ the prompts provided in the CLIP GitHub repository\footnote{\url{https://github.com/openai/CLIP}.}. Our experiments indicated that the inclusion of these prompts enhanced the accuracy of our model by $1-2\%$.

\begin{table}[h]
  \caption{Evaluation of the
  Classification Accuracy vs. Privacy of \dpclip (Average over 10 trials). More details about the standard deviations can be found in Table \ref{tab:fullutilityprivacytradeoff}}
  \label{tab1:utilityprivacytradeoff}
  \centering
  \scalebox{0.9}{
  \begin{tabular}{llllll}
    \toprule
     & MNIST & Fashion-MNIST & CIFAR-10 & SVHN\\
    \midrule
    $\epsilon=\infty$ & 98.97 & 92.08 & 95.07 & 93.81 \\
    $\epsilon=10$ & 98.70 & 91.42 & 95.48 & 92.53 \\ 
    $\epsilon=3$ & 98.67 & 91.05 & 95.16 & 91.56 \\ 
    $\epsilon=1$ & 98.40 & 90.35 & 94.74 & 90.31 \\ 
    $\epsilon=0.5$ & 98.12 & 89.76 & 94.03 & 88.92 \\ 
    $\epsilon=0.25$ & 97.66 & 88.83 & 93.19 & 87.18 \\ 
    zero-shot & 48.06 & 66.31 & 88.31 & 24.86 \\
    \bottomrule
  \end{tabular}
  }
\end{table}

We present the classification accuracy of \dpclip on the four datasets under various $\epsilon$ in Table \ref{tab1:utilityprivacytradeoff}. We report the mean of 10 trials in Table \ref{tab1:utilityprivacytradeoff} and leave the standard deviations to Appendix \ref{sec.classification}. We observe that \dpclip is able to recover features under the regime with stringent privacy parameters. In particular, for all datasets, $\epsilon=1$ performs within $1\%$ of $\epsilon=10$, which indicates the strong potential of \dpclip to offer better privacy guarantees while maintaining utility. 

To further evaluate the performance of \dpclip, we compare our \dpclip\ against other differentially private methods that achieve state-of-the-art results on these datasets. We consider DP-KIP \citep{dpkip}, DP-SGD with Tempered Sigmoid (DP-SGD (TS) for short) \citep{temperedsigmoid}, Private-kNN \citep{privateknn}, and Active Learning \citep{activelearning}, and DP-SGD on over-parameterized models (DP-SGD (large) for short) \cite{de2022unlocking}. We present the comparisons in Table \ref{table:all} below. For brevity, only the best results from each paper are included. 
\begin{table}[tph]
  \centering
  \caption{Comparison with state-of-the-art DP methods on MNIST, FashionMNIST, CIFAR-10 and SVHN, with varying parameter $\epsilon$. 
  }
  \scalebox{0.72}{
  \begin{tabular}{p{2.3cm}cccp{2.3cm}cc}
    \multicolumn{3}{c}{MNIST} & & \multicolumn{3}{c}{Fashion-MNIST} \\
    \cmidrule(r){1-3} \cmidrule(l){5-7} 
    & $\epsilon$ & Accuracy & \qquad & & $\epsilon$ & Accuracy\\
    \cmidrule(r){1-3} \cmidrule(l){5-7} 
    DP-KIP  & 10 & 97.96 &\qquad & DP-KIP  & 10 & 90.2 \\
    \dpclip & 10 & \textbf{98.70}  &\qquad & \dpclip & 10 & \textbf{91.42} \\
    \cmidrule(r){1-3} \cmidrule(l){5-7} 
    Active Learning & 3 & 97.3 &\qquad &  &  &  \\
    \dpclip & 3 & \textbf{98.67} &\qquad &  &  &  \\
    \cmidrule(r){1-3} \cmidrule(l){5-7}
    DP-SGD (TS)  & 2.93 & 98.1  &\qquad & DP-SGD (TS)  & 2.7 & 86.1 \\
    DP-KIP  & 1 & 97.78  &\qquad & DP-KIP  & 1 & 88.3 \\
    \dpclip & 1 & \textbf{98.40}  &\qquad & \dpclip & 1 & \textbf{90.35} \\
    \cmidrule(r){1-3} \cmidrule(l){5-7} 
    Private-kNN  & 0.47 & \textbf{98.8} &  &  &  \\
    \dpclip & 0.5 & 98.12  &  &  &  \\
    \cmidrule(r){1-3}
    & & & & & & \\
    \multicolumn{3}{c}{CIFAR-10} & & \multicolumn{3}{c}{SVHN}\\
    \cmidrule(r){1-3} \cmidrule(l){5-7}
    & $\epsilon$ & Accuracy & \qquad & & $\epsilon$ & Accuracy\\
    \cmidrule(r){1-3} \cmidrule(l){5-7} 
    DP-SGD (large) & 4 & \textbf{96.1} &\qquad & Active Learning  & 6 & 85.0 \\
    \dpclip & 3 & 95.16 &\qquad & \dpclip & 3 & \textbf{91.56} \\
    \cmidrule(r){1-3} \cmidrule(l){5-7}
    DP-SGD (large)  & 1 & 94.7 &\qquad & Private-kNN  & 0.49 & \textbf{91.6} \\
    \dpclip & 1 & \textbf{94.74} &\qquad & \dpclip & 0.5 & 88.92 \\
     \cmidrule(r){1-3} \cmidrule(l){5-7}
  \end{tabular}
  \label{table:all}
  }
\end{table}

From Table \ref{table:all}, we can see that for most cases, \dpclip outperforms all other methods on all four datasets when considering a smaller or equal $\epsilon$.  
This accuracy improvement is by leveraging both pretraining and using extra caption data for \dpclip , which is not present for DP-SGD.
Although the performance of Private-kNN exceeds \dpclip\ on the MNIST and SVHN datasets when $\epsilon < 0.5$, our \dpclip\ offers more flexibility. Unlike Private-kNN, which is limited to classification tasks, \dpclip can be applied to
a variety of more complex downstream tasks.

To capture a more realistic scenario, we employ \dpclip on the German Traffic Sign Recognition Benchmark (GTSRB) dataset \cite{gtsrbdataset}, characterized by a distribution shift relative to the pre-training dataset, and CIFAR-100 \cite{cifar100dataset}, where the number of training samples per class is relatively low. We conduct 10 trials for each experiment, presenting the average accuracy along with the standard error in Table \ref{tab3:cifar100gtrsb}. Our CIFAR-100 results demonstrate comparability to the state-of-the-art (SOTA) findings in \cite{de2022unlocking}. Notably, our utility loss remains comparatively milder when $\epsilon \le 1$. Despite the distribution shift in GTSRB, our results exhibit robust performance.

\begin{table}[ht]
  \caption{Evaluation on datasets with distribution shifts (CIFAR-100 and GTSRB)}
  \label{tab3:cifar100gtrsb}
  \centering
  \scalebox{0.95}{
  \begin{tabular}{cccc}
    \toprule
    $\epsilon$ & CIFAR-100 & GTSRB \\
    \midrule
    $10$ & 79.64 $\pm$ 2.25e-03 & 92.47 $\pm$ 2.50e-03 \\ 
    $3$ & 78.22 $\pm$ 2.57e-03 & 91.31 $\pm$ 5.90e-03 \\ 
    $1$ & 76.29 $\pm$ 3.37e-03 & 88.59 $\pm$ 2.85e-03 \\ 
    $0.5$ & 74.62 $\pm$ 3.05e-03 & 84.95 $\pm$ 6.93e-03 \\ 
    $0.25$ & 71.55 $\pm$ 3.16e-03 & 77.55 $\pm$ 7.64e-03 \\ 
    0 (zero-shot) & 62.36 $\pm$ 0.00 & 27.59 $\pm$ 0.00 \\
    \bottomrule
  \end{tabular}
  }
\end{table}


We also consider the low data setting following the framework outlined in \cite{pinto2023pillar}. In this experiment, we only use a subset of the GTSRB training data and compare \dpclip with their method PILLAR \cite{pinto2023pillar} under the same privacy regime and the same training data percentage.

\begin{table}[ht]
  \caption{Evaluation under low data settings on GTSRB}
  \label{tab4:gtsrb comparison}
  \centering
  \scalebox{0.9}{
  \begin{tabular}{cccc}
    \toprule
     $\epsilon$ and Training Percentage Used & \dpclip & PILLAR \\
    \midrule
    $\epsilon=0.1$ at 10\% Train Data & 28.33 $\pm$ 4.28e-03 & $\sim$ 8 \\ 
    $\epsilon=0.1$ at 50\% Train Data & 45.43 $\pm$ 1.42e-02 & $\sim$ 33 \\ 
    $\epsilon=0.7$ at 10\% Train Data & 37.98 $\pm$ 1.04e-02 & $\sim$ 38 \\ 
    $\epsilon=0.7$ at 50\% Train Data & 79.42 $\pm$ 6.48e-03 & $\sim$ 59 \\ 
    \bottomrule
  \end{tabular}
  }
\end{table}

We observe the expected trends where higher $\epsilon$ and higher percentages of train data lead to better performance for both \dpclip and PILLAR. For all four settings, \dpclip is able to either match or outperform PILLAR in terms of accuracy, which demonstrates that \dpclip can achieve strong performance on an out-of-distribution dataset in low data settings.

\subsection{Image Captioning Results}
\label{sec:imgcap}
The {\em Bootstrapping Language Image Pre-training} (BLIP) \citet{li2022blip} is a novel framework for VLP (Vision-Language Pre-training \citet{vip}) that offers broad applicability to various downstream tasks and there have seen growing interest in image captioning because of its potential to aid the blind community \citep{vizwiz} recent years.  Hence, we adopt BLIP for this task, training the private visual and textual representations jointly using caption-image pairs. 
Prior to this work, differential privacy has not been applied to image captioning. 
We establish the first baselines on differentially private image captioning tasks.

Our goal is to demonstrate the utility and privacy of learned private representations rather than a powerful differentially private vision-language model. Hence, instead of fine-tuning BLIP, we only use contrastive loss while training, aiming to learn the DP representations. More specifically, we train BLIP privately on a subset ($15\%$) of the train set of Vizwiz Image Captioning dataset \citep{vizwiz} and evaluate it on a subset ($15\%$) of the test set using contrastive loss. We present several common evaluation metrics for various privacy parameters: $\epsilon= 0$ (zero-shot), 0.5, 1, 5, and $\infty$ (non-private).

Table \ref{tbl: comparison} compares the performances on image captioning accuracy for different privacy-preserving levels, including the zero-shot, private (of vairous level), and non-private results. We observe that as $\epsilon$ increases, implying weaker privacy, the performances across all metrics improve. This is in line with the privacy-utility trade-off: as one weaken privacy guarantees, one typically expect to see better utility. We also note that the decrease in performances is not uniform across all metrics: some metrics, such as CIDEr, are more sensitive to changes in $\epsilon$ than others.

\begin{table}[ht]
  \caption{Evaluation of the Accuracy vs. Privacy of \dpblip on Image Captioning}
  \label{tbl: comparison}
  \centering
  \scalebox{0.8}{
  \begin{tabular}{cccccc}
    \toprule
     & $\epsilon=0$ & $\epsilon=0.5$ & $\epsilon=1$ & $\epsilon=5$  & $\epsilon=\infty$\\
     & (zero-shot) & & & & (non-private)\\
    \midrule
    Bleu\_1 & 0.587	& 0.659	& 0.665 & 0.668 & 0.679 \\
    Bleu\_2 & 0.406	& 0.475 & 0.478	& 0.484	& 0.492 \\
    Bleu\_3 & 0.270	& 0.333 & 0.333	& 0.338	& 0.343 \\
    Bleu\_4 & 0.171	& 0.222 & 0.222	& 0.230	& 0.231 \\
    ROUGE\_L & 0.417 & 0.454 & 0.457 & 0.475 & 0.463 \\
    CIDEr & 0.431 & 0.584 & 0.584 & 0.690 & 0.695 \\ 
    SPICE & 0.165 & 0.189 & 0.193 & 0.210 & 0.207 \\
    \bottomrule
  \end{tabular}
  }
\end{table}

\subsection{Visual Question Answering Results}
We also repeat similar experiments on the VQA task, but instead of framing it as a multi-answer classification problem \citet{10.1007/978-3-030-58577-8_7, li2020oscar}, we formulate it as an answer generation task \citep{li2021align, li2022blip}. This formulation allows for open-ended VQA, where the model generates answers rather than selecting from a predefined set of options, which is consistent with the task used in \citet{li2022blip} and \citet{li2021align}. As depicted in Figure \ref{fig:clip-pretraining-blip_vqa} (right) in Section \ref{app.prelim}, during the training process, an image-question pair is encoded into multimodal embeddings, which are then fed into an answer decoder. We use contrastive loss on the question-answer and image pairs to train private representation and then without further finetuning, we evaluate our results using the exact-match accuracy metric.

The goal of the experiment is to demonstrate that adding noise to achieve DP does not significantly impact the accuracy of the results. The objective is not to train the model to compete with non-private state-of-the-art results, but rather to showcase that our approach achieves a comparable level of accuracy without compromising too much privacy.

We used the exact-match accuracy \ref{sec:VQA} and the \href{https://visualqa.org/evaluation.html}{top-three accuracy}, and $\min\{1, \text{\# humans saying that answer}/3\}$, for evaluations. This experiment reveals a privacy-utility trade-off, which is consistent with previous experiments. Due to the page limits, we defer the detailed experiments to the appendix.

\section{Theoretical Analysis of \dpclip}\label{sec: theory}

In this section, we analyze the feature learning capacity of DP-CLIP and derive the privacy-utility trade-off under the linear representation and loss setting. Such a simplified setting has been commonly used in the deep learning theory literature to shed light on understanding complicated deep learning phenomena. For example, the linearized loss function for analyzing representation learning has been used in metric learning \citep{schroff2015facenet,he2018triplet}, contrastive learning \citep{ji2021power} and multimodal contrastive learning \citep{won2021multimodal,alsan2021multimodal,nakada2023understanding}.
The linear representation setting has been widely adopted in transfer learning and self-supervised learning \citep{jing2021understanding,tian2021understanding,ji2021power,wu2022sparse,tian2022deep,nakada2023understanding}.

Concretely, suppose that we observe $n$ pairs of data $\{(x_i, \tilde x_i)\}_{i=1}^n \subset \R^{d_1}\times \R^{d_2}$.
Let $r$ be the dimension of the representation space ($r < d$).
We train dual \textit{linear representations} $f(x) = G_1 x$ and $\tilde f(x) = G_2 x$, where $G_1 \in \R^{r \times d_1}$ and $G_2 \in \R^{r \times d_2}$, simultaneously with the following contrastive linear loss through noisy gradient descent. 
For notational brevity, let $G \triangleq [G_1, G_2] \in \R^{r \times d}$, where $d \triangleq d_1 + d_2$.

We aim to obtain $G_1$ and $G_2$ that minimize the following linearized loss function:
\begin{align}
    \mL(G_1, G_2) &= - \frac{1}{n} \sum_i \langle G_1 x_i, G_2 \tilde x_i \rangle\nonumber\\
    &\quad+ \frac{1}{n(n-1)} \sum_{i \neq j} \langle G_1 x_i, G_2 \tilde x_j \rangle + \Pi(G),\label{loss: linear}
\end{align}
where the penalty term $\Pi(G) \triangleq (\alpha/4) \|G G^\top - I\|_F^2$ with $\alpha > 0$ is added to normalize $G$.
Note that the CLIP loss \ref{loss: CLIP} becomes equivalent to the loss in \eqref{loss: linear} without penalty when $\tau \to \infty$. 

For observed data $\{(x_i, \tilde x_i)\}_{i=1}^n \subset \R^{d_1 + d_2}$, we consider the following spiked covariance model \citep{johnstone2001distribution,bai2012sample,yao2015sample,zhang2018heteroskedastic,zeng2019double,ji2021power,nakada2023understanding} as the data generation process.
\begin{align}
    x_i &= U_1^* z_i + \xi_i,\ \ \tilde x_i = U_2^* z_i + \tilde \xi_i.\label{model: dual spiked covariance model}
\end{align}
where $U_1^*$ and $U_2^*$ are $d_1\times r$ and $d_2 \times r$ orthogonal matrces, respectively.
Since the model \eqref{model: dual spiked covariance model} is only identifiable up to rotation, we assume that $\Sigma_z$ is a diagonal matrix. Without loss of generality, we further assume that $\|\Sigma_z\| = 1$. We assume that $z_i$, $\xi_i$ and $\tilde\xi_i$ are mean $0$ sub-Gaussian random variables with parameters bounded by a universal constant. Furthermore, we assume the independence of variables; $z_i \indep \xi_i$, $\tilde z_i \indep \tilde \xi_i$, and $\xi_i \indep \tilde \xi_i$.

There have been several works on analyzing the convergence of DP-SGD \citep{abadi2016deep} with per-sample clipping \citep{yu2019gradient,bassily2019private,feldman2020private,chen2020understanding,yang2022normalized,bu2022automatic,fangimproved}.
Closely related works are \citet{yang2022normalized} and \citet{fangimproved}. 
However, the contrastive loss function of CLIP cannot be decomposed into the sum of per-sample losses, since CLIP learns by contrasting modalities across sampled pairs. 
This is the reason why \dpclip employs per-\textit{mini-batch} clipping.
In addition, in spite of the linearization, our loss function \eqref{loss: linear}, like the original CLIP loss, is neither convex nor globally Lipschitz, which makes the theoretical analysis highly nontrivial.

\subsection{Privacy-utility Trade-off of \dpclip}

Let $G_1^*, G_2^*$ be the minimizer of the population loss $\mathbb{E}[\mL(G)]$. For simplicity, we assume the regularization parameter $\alpha = \Theta(1)$ is of constant order.
Since pretrained encoder is often used in downstream tasks, where the output of the encoder is fed into neural networks or linear probes, the essential information of the learned representation is contained in the linear transformation. For this reason, we measure the performance of the learned representations through the excess loss of information defined as $\min_{A \in \R^{r \times r}} \|A G_1 - G_1^*\|_F \vee \min_{A \in \R^{r \times r}} \|A G_2 - G_2^*\|_F$. 
For ``good" encoders, we expect that a certain linear transformation of it is close to the representations obtained using infinite number of training samples.

Before presenting our results, we introduce notations.
For two sequences of positive numbers $\{a_k\}_k$ and $\{b_k\}_k$, we write $a_k\lesssim b_k$ if and only if there exists a constant $C > 0$, independent of the index $k$, such that $\sup_{k \in \mathcal{K}} (a_k/b_k) < C$.
Moreover, we write $a_k \ll b_k$ when $\sup_{k \in \mathcal{K}} (a_k/b_k) \leq C_u$ holds for a sufficiently large universal constant $C_u > 0$ common throughout the paper.
For any matrix $A$, we denote $\|A\|$ and $\|A\|_F$ as the operator norm and Frobenius norm of $A$ respectively. 
For any matrix $A$, let $\lambda_{\min}(A)$ and $\lambda_{\max}(A)$ be the minimum and maximum singular values of $A$, respectively.
For any zero-mean random variable $X$, we define its covariance matrix as $\Sigma_X \triangleq \mathbb{E}[X X^\top]$.
Let the signal-to-noise ratio for $x$ and $\tilde x$ be $s_1^2 \triangleq \|\Sigma_z\|/ \|\Sigma_\xi\|$ and $s_2^2 \triangleq \|\Sigma_z\|/\|\Sigma_{\tilde \xi}\|$, respectively.
\begin{assumption}\label{asm: regime}
    Assume that $d > r$ and
        $n \gg r(r + s_1^{-2} r_e(\Sigma_\xi) + s_2^{-2} r_e(\Sigma_{\tilde \xi}))^2 \log^3(T(n+d))$,
    where $r_e$ is the effective rank defined as $r_e(A) \triangleq \Tr(A)/\|A\|$ for any square matrix $A$. 
\end{assumption}

\begin{assumption}[Signal-to-Noise Ratio]\label{asm: snr}
    Assume that $\min\{s_1^2, \ s_2^2\} \gtrsim 1$.
\end{assumption}

\begin{assumption}[Signal Condition Number]\label{asm: signal condition number}
    Assume that $\kappa \triangleq \lambda_{\max}(\Sigma_z)/\lambda_{\min}(\Sigma_z) \lesssim 1$.
\end{assumption}

Assumption \ref{asm: regime} ensures that we have an effective number of samples to separate the core signal from the noise.
Assumption \ref{asm: snr} is a mild condition on the signal-to-noise ratio. It allows the noise to be the same strength as signal.
Assumption \ref{asm: signal condition number} ensures that core features are strongly shared between the two modalities.

Here we introduce the privacy-utility trade-off of \dpclip\ under linear loss. 
\begin{theorem}[Privacy-utility Trade-off]\label{thm: privacy utility tradeoff}
    Suppose Assumptions \ref{asm: regime}, \ref{asm: snr} and \ref{asm: signal condition number} hold.
    Assume that $\alpha = \Theta(1)$.
    Let $G^{(T)}$ be the representation obtained from Algorithm \ref{alg:DP-GD} with loss $\mL$.
    Suppose that the initial representation $G^{(0)}$ satisfies
    $
        \min_{O \in \R^{r \times r}: O^\top O = I} \|O G^{(0)} - \hat G\|_F \ll 1.
    $    
    Choose $c \gg 1$, $b = \ceil{\nu n}$, where $\nu \in (0, 1)$ is a constant.
    Also choose $\eta > 0$ and $\sigma > 0$ as
    \begin{align*}
    \eta &= \{\sigma \sqrt{T(rd + \log(T(n+d)))}\}^{-1},\\
    \sigma &= C_\sigma \sqrt{T \log(1/\delta)}/(n \epsilon),
    \end{align*}
    where $C_\sigma$ is a constant appearing in Proposition \ref{prop: dp}.
    Then, Algorithm \ref{alg:DP-GD} under loss $\mL$ is $(\epsilon, \delta)$-DP and for sufficiently large $T$,
    \begin{align}
        &\min_{A \in \R^{r \times r}} \|A G_1^{(T)} - G_1^*\|_F \vee \min_{A \in \R^{r \times r}} \|A G_2^{(T)} - G_2^*\|_F\nonumber\\
        \lesssim & \underbrace{\exp(-\frac{n\epsilon }{8 \kappa C_\sigma \sqrt{\log(1/\delta) \{rd + \log(T(n+d))}\}})}_{\text{optimization error}}\nonumber\\
        &+ \underbrace{\frac{\log^{1/4}(1/\delta) \{rd + \log(T(n+d))\}^{1/4}}{\sqrt{n\epsilon}}}_{\text{the cost of privacy}}\nonumber\\
        &+ \underbrace{\sqrt{\frac{r(r + s_1^{-2} r_e(\Sigma_\xi) + s_2^{-2} r_e(\Sigma_{\tilde \xi}))^2 \log^3(n+d)}{n}}}_{\text{statistical error}}.\label{eq: tradeoff}
    \end{align}
    
    holds with probability at least $1 - O((n+d)^{-1})$.
\end{theorem}

\paragraph{Proof Outline of Theorem~\ref{thm: privacy utility tradeoff}.}
This result follows from the linear convergence result (Theorem \ref{thm: convergence to global minimum}) of Algorithm \ref{alg:DP-GD};
we can bound the distance between $G^{(T)}$ and the global minimizer of the loss $\mL$ by three components, a linear converging term, the error from the injected noise, and the error from subsampling.
To show this result, we first derive the one-step linear convergence bound for non-stochastic gradient descent without noise injection. To this end, we use the fact that $\mL$ is locally strongly convex and directionally smooth around its global minimum.
For noisy stochastic gradient descent, we need to control the accumulation of errors coming from both privacy noise and subsampling.
For this purpose, we exploit the fact that $\pG\mL(G^{(t)}; \mB^{(t)})$, a mini-batch gradient of $\mL$ evaluated at $G^{(t)}$, is an unbiased estimator of $\pG\mL(G^{(t)})$, and control the deviation of the accumulated errors through the martingale concentration inequality. 
The accumulated error from subsampling is controlled by the Bernstein concentration bound from \citet{bardenet2015concentration}, which turns out to be negligible since the batch size is chosen to be proportional to the number of samples.
Given the linear convergence result, we set the value of $\eta$ and $\sigma$ as specified in Proposition \ref{prop: dp} to conclude the proof.
The proof and more detailed statement of Theorem~\ref{thm: privacy utility tradeoff} that specifies the exact condition on $T$ is available in Corollary \ref{cor: convergence to population global minimum 2 ap} in the appendix.

In \eqref{eq: tradeoff}, the right-hand side consists of three terms: optimization error, privacy cost, and statistical error.
The optimization error decreases exponentially in $n$, since the loss function $\mL$ behaves well locally around the global minimum. This term grows with $T$ because the algorithm runs in two stages: in the first stage, the error decreases exponentially in $T$ (see details in Theorem \ref{thm: convergence to global minimum}), and in the second stage, when $G^{(T)}$ reaches a certain stable region, the optimization error starts to increase (slowly) if we continue to run more gradient descent updates. 
We also note that the optimization error is dominated by the term for the cost of privacy when $n\epsilon/\sqrt{\log(1/\delta)}$ is large. 
The second term corresponds to the additional cost to preserve privacy. Ignoring the logarithmic term $\log(T(n+d))$, the cost increases proportionally to $\log^{1/4}(1/\delta)/\epsilon^{1/2}$.
This rate also appears similarly in \citet{chen2020understanding,yang2022normalized,fangimproved} for the analysis of DP-SGD. Our technical analysis differs from theirs as they consider per-sample clipping for loss functions that satisfy certain smoothness condition, which does not hold for our loss \ref{loss: linear}. 
We also note that the loss \ref{loss: linear} is not decomposable to apply per-sample clipping.
Also, similar privacy cost bounds appear in the convergence analysis of differentially private gradient descent and related algorithms \citep{wang2017differentially,wang2019differentially,zhang2019gradient,wang2020differentially,cai2021cost,cai2023score}.
The statistical error term is due to the irreducible error coming from finite samples. 
The term depends on $r + s_1^{-2} r_e(\Sigma_\xi) + s_2^{-2} r_e(\Sigma_{\tilde \xi})$, which is trivially bounded by $O(r + d)$ under Assumption \ref{asm: snr}. When either the effective rank of noise covariance is small or the signal-to-noise ratio is large, the statistical error term becomes small.

\begin{remark}
    In the analysis, the intrinsic dimension $r$ of the input data is assumed to be known. In practical situation, we can either choose $r$ based on certain metric such as cross validation in downstream tasks, or estimate $r$ based on the spectral decay of the cross-covariance matrix.
\end{remark}

\begin{remark}
    The initial value condition is satisfied if $\min_{O : O^\top O = I} \|O G^{(0)} - \hat G\| \ll 1/\sqrt{r}$.
    Thus the condition is considered to be weak when $r$ is small.
    As in our experiments in Section \ref{sec:exp}, we can employ initial representations trained with non-private optimizers with certain number of samples.
\end{remark}

\section{Discussion}\label{sec: discussion}

In this paper, we introduce \dpclip, a novel approach that integrates differential privacy into the CLIP to address privacy concerns associated with vision-language tasks. To our knowledge, this is the first attempt to apply differential privacy approaches to multimodal training, where we have created a framework applicable to a variety of vision-image tasks that have not previously been explored in the DP literature.
We conduct extensive experiments to demonstrate the effectiveness of our \dpclip\ on three tasks: image classification, image captioning and visual question answering tasks. 
In addition, we theoretically prove the convergence of our algorithm under linear representation settings, and present a privacy-utility trade-off for \dpclip in the situation where gradients are clipped per batch and the loss function does not satisfy the smoothness conditions such as Lipschitz smoothness.


Our work identifies several areas for further investigation. Future work includes conducting experiments on privatizing other multimodal models and evaluating them across a broader range of vision-language downstream tasks, such as visual entailment or image-text retrieval \citep{li2021align}. Although previous work has suggested that DP can protect against common privacy attacks such as reconstruction attacks and membership inference attacks \citep{membership-attacks}, investigating the empirical privacy auditing on \dpclip\ is also interesting. Additionally, while our theoretical analysis focuses on linear representation functions, non-linear representations need further exploration. 
Moreover, \citet{vyas2023provable} introduced a copyright protection framework for generative models, which relates to differential privacy as a mathematical measurement. Given that our method learns differentially private image representations, it has the potential to generate copyright-protected images. 


\bibliography{ICML/icml_2024}
\bibliographystyle{icml2024}

\newpage
\appendix
\onecolumn

\section{Additional Preliminaries on CLIP}\label{app.prelim}

\begin{figure}[!h]
    \centering
    \includegraphics[width=.3\linewidth]{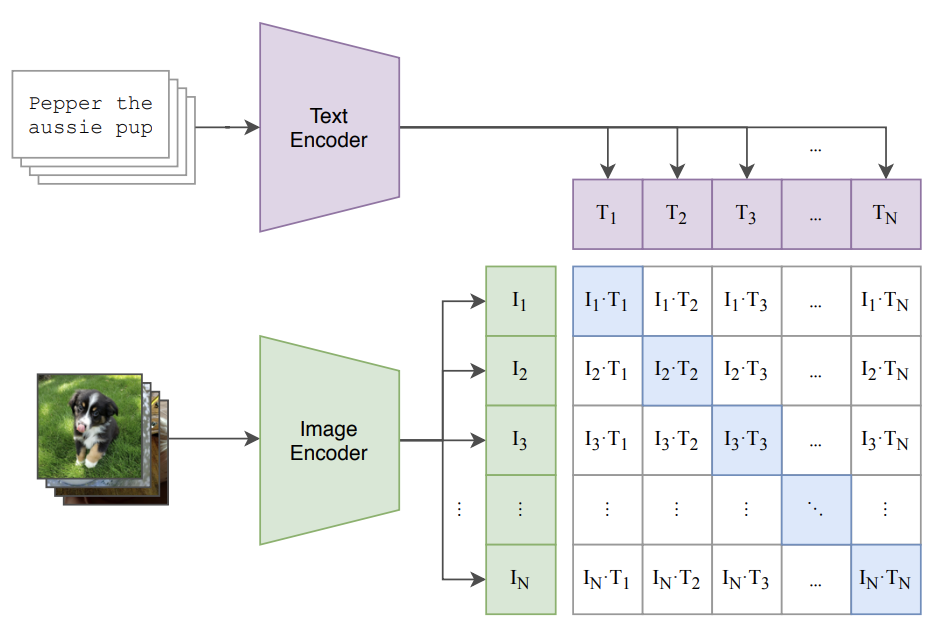}
    \qquad \qquad
    \includegraphics[width=.4\linewidth]{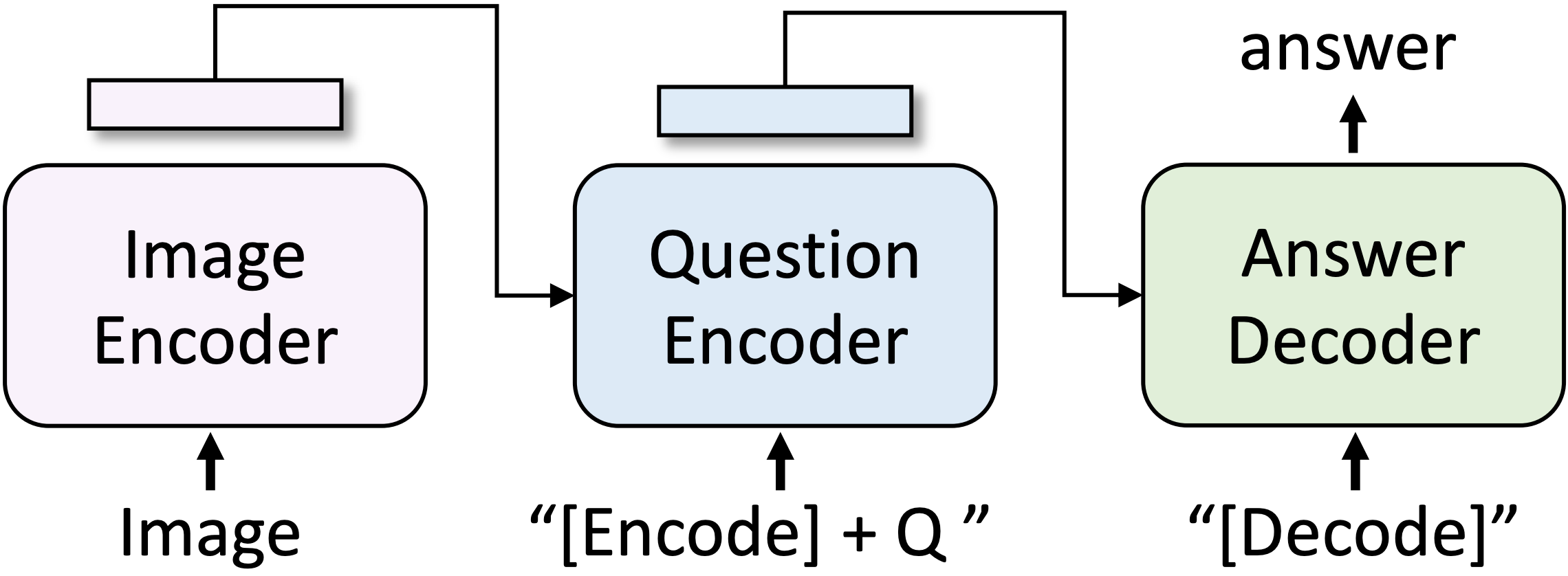}
    \caption{CLIP Pretraining Process from \citet{radford2021learning} (Left) and Structure of BLIP for VQA from \citet{li2022blip} (Right)}
    \label{fig:clip-pretraining-blip_vqa}
\end{figure}

The goal of CLIP is to train an image encoder and a text encoder, by maximizing the cosine similarity of correct image-text pairs (highlighted entries on the diagonal in Fig. \ref{fig:clip-pretraining-blip_vqa}) and minimizing the cosine similarity of incorrect image-text pairs (other non-diagonal entries in Fig. \ref{fig:clip-pretraining-blip_vqa}.)

\section{Detailed Experimental Results}\label{app:exp}
In this section, we provide additional experimental setups and results. 
\subsection{Image Classification}\label{app.classification}

We present the hyperparameters used in image classification task in Table  \ref{tab4:hyperparams}. 

\begin{table}[ht]
  \caption{Tuned Hyperparameters for Image Classification \dpclip}
  \label{tab4:hyperparams}
  \centering
  \begin{tabular}{llllll}
    \toprule
     & MNIST & Fashion-MNIST & CIFAR-10 & SVHN\\
     
    \midrule
    lr & 1e-05  & 1e-05  & 1e-05 & 1e-05 \\ 
    betas & (0.9, 0.98) & (0.9, 0.98) & (0.9, 0.98) & (0.9, 0.98) \\ 
    eps & 1e-06 & 1e-06 & 1e-06 & 1e-06 \\ 
    weight decay & 0.01 & 1e-06 & 1e-06 & 1e-06 \\ 
    num epochs & 30 & 30 & 30 & 15 \\ 
    batch size & 32 & 32 & 32 & 32 \\
    \bottomrule
  \end{tabular}
\end{table}

We also present the full image classification results with both averages and standard deviations over 10 trials in Table \ref{tab:fullutilityprivacytradeoff}.
\begin{table}[ht]
  \caption{Evaluation of the
  Classification Accuracy vs. Privacy of \dpclip (Average and StdDev)}
  \label{tab:fullutilityprivacytradeoff}
  \centering
  \scalebox{1}{
  \begin{tabular}{llllll}
    \toprule
     & MNIST & Fashion-MNIST & CIFAR-10 & SVHN\\
    \midrule
    $\epsilon=\infty$ & 98.97 $\pm$ 0.00 & 92.08 $\pm$ 2.55e-03 & 95.07 $\pm$ 0.00 & 93.81 $\pm$ 1.65e-03 \\
    $\epsilon=10$ & 98.70 $\pm$ 6.45e-04 & 91.42 $\pm$ 1.56e-03 & 95.48 $\pm$ 1.18e-03 & 92.53 $\pm$ 1.16e-03 \\ 
    $\epsilon=3$ & 98.67 $\pm$ 7.34e-04 & 91.05 $\pm$ 1.76e-03 & 95.16 $\pm$ 1.20e-03 & 91.56 $\pm$ 1.79e-03 \\ 
    $\epsilon=1$ & 98.40 $\pm$ 1.28e-03 & 90.35 $\pm$ 1.44e-03 & 94.74 $\pm$ 1.19e-03 & 90.31 $\pm$ 2.59e-03 \\ 
    $\epsilon=0.5$ & 98.12 $\pm$ 1.23e-03 & 89.76 $\pm$ 2.28e-03 & 94.03 $\pm$ 1.36e-03 & 88.92 $\pm$ 3.14e-03 \\ 
    $\epsilon=0.25$ & 97.66 $\pm$ 6.18e-04 & 88.83 $\pm$ 1.35e-03 & 93.19 $\pm$ 3.67e-03 & 87.18 $\pm$ 2.93e-03 \\ 
    Zeroshot & 48.06 $\pm$ 0.00 & 66.31 $\pm$ 0.00 & 88.31 $\pm$ 0.00 & 24.86 $\pm$ 0.00 \\
    \bottomrule
  \end{tabular}
  }
\end{table}

\subsection{VQA}\label{sec:VQA}

The {\em Bootstrapping Language Image Pre-training} (BLIP) \citet{li2022blip} is a novel framework for VLP (Vision-Language Pre-training \citet{vip}) that offers broad applicability to various downstream tasks. It introduces (a) \emph{Captioning and Filtering} (CapFilt), a pioneering method for dataset bootstrapping that enables learning from noisy image-text pairs, and (b)  \emph{Multimodal mixture of Encoder-Decoder} (MED), which is a novel model architecture capable of functioning as a unimodal encoder, an image-grounded text encoder, or an image-grounded text decoder. This versatility facilitates effective multi-task pre-training and flexible transfer learning. The MED model is jointly pretrained using three vision-language objectives: \emph{ image-text contrastive learning} \citep{radford2021learning} to align the vision and language representations, \emph{image-text matching} \citep{Image_Text_Matching}  to distinguish between positive and negative image-text pairs, and \emph{image-conditioned language modeling}\citep{pmlr-v32-kiros14} to generate good textual descriptions given an image. We decided to use BLIP as our backbone since it currently achieves state-of-the-art performance on the image captioning task.

Visual Question Answering (VQA) has been increasingly used in many fields, such as healthcare, education, and social media \citep{srivastava2021visual}. In Visual Question Answering (VQA), an image (I) is provided along with a related question (Q) in natural language form, and the goal is to generate accurate and meaningful answers (A). 
However, the images and text data used in training may contain sensitive information, and it is of paramount importance to ensure privacy preservation measures are in place \citep{bara2022privacy}. For similar reasons as the image-captioning section in 4.2, because BLIP shows such strong performance on VQA tasks, we decided to use it as the backbone for our VQA experiments.

For VQA tasks, instead of framing it as a multi-answer classification problem \citet{10.1007/978-3-030-58577-8_7, li2020oscar}, BLIP takes a different approach by formulating it as an answer generation task \citep{li2021align, li2022blip}. This formulation allows for open-ended VQA, where the model generates answers rather than selecting from a predefined set of options, which is consistent with the task used in \citet{li2022blip} and \citet{li2021align}. As depicted in Figure \ref{fig:clip-pretraining-blip_vqa}, during the training process, an image-question pair is encoded into multimodal embeddings, which are then fed into an answer decoder. We use contrastive loss on the question-answer and image pairs to train private representation and then without further finetuning, we evaluate our results using the exact-match accuracy metric.

The goal of the experiment in this section is to demonstrate that adding DP noise does not significantly impact the accuracy of the results, compared to the non-private method, when all other parameters are held constant. The objective is not to train the model to compete with non-private state-of-the-art results, but rather to showcase that our approach achieves a comparable level of accuracy without compromising privacy. We note that BLIP can achieve state-of-the-art results on VQA with a much higher accuracy rate of $78.25 \%$ \citep{li2022blip}. However, in our study, we deliberately refrain from extensive parameter optimization and instead focus on providing a baseline analysis. As a result, we report lower accuracy results compared to the fine-tuned BLIP approach. From Table \ref{tab:blipperformance}, we can see that our model maintains utility even as privacy measures are increased, suggesting its resilience to noise.

\begin{table}[h!]
  \caption{Evaluation on the abstract scene VQA2.0 dataset using exact-match accuracy}
  \label{tab:blipperformance}
  \centering
  \begin{tabular}{ccccc}
    \toprule
    $\epsilon=0.5$ & $\epsilon=1$ & $\epsilon=3$ & $\epsilon=10$ & $\epsilon=\infty$ (non-private)\\
    \midrule
    52.94\% & 52.94\% & 52.94\% & 52.95\% & 55.22\%\\
    \bottomrule
  \end{tabular}
\end{table}

We report the exact-match accuracy mentioned in \ref{tab:blipperformance}, here, we additionally evaluate it using another metric, top-three accuracy, that is robust to iter-human variability in phrasing the answers, introduced in \url{https://visualqa.org/evaluation.html}. The evaluation code was taken from ALBEF’s github repository, where we consider the top three answers given by humans to a question and our accuracy is taken to be $\min\{1, \text{\# humans saying that answer}/3\}$ and we output the average accuracy over the training set. The following results, shown in Tab. \ref{tab:blipvqamodified}, suggests our privacy-aware approach can achieve comparable performance to the non-private method.


\begin{table}[h!]
  \caption{Evaluation on the abstract scene VQA2.0 dataset using top-three accuracy}
  \label{tab:blipvqamodified}
  \centering
  \begin{tabular}{cccccc}
    \toprule
     $\epsilon=0.0001$ & $\epsilon=0.5$ & $\epsilon=5$ & $\epsilon=50$\\
     
    \midrule
    0.4686 & 0.4717 & 0.4784 & 0.4924 \\ 

    \bottomrule
  \end{tabular}
\end{table}

\section{Proofs of Theoretical Results}

Before going into the proofs, we introduce notations to be used in later sections.

\subsection{Notation}
In this section, we introduce notations to be used.
We write $a_k = O(b_k)$ if $a_k \lesssim b_k$ holds and $a_k = \Omega(b_k)$ if $a_k \gtrsim b_k$ holds. $\mathbb{O}_{d,r} \triangleq \{O \in \R^{r\times d} : O^\top O = I_r\}$ as a set of orthogonal matrices of order $d \times r$.
We write $a \vee b$ and $a \wedge b$ to denote $\max(a, b)$ and $\min(a, b)$, respectively.
For any matrix $A$, let $\lambda_j(A)$ be the $j$-th largest singular value of $A$. Let $\lambda_{\min}(A)$ and $\lambda_{\max}(A)$ be the minimum and maximum singular values of $A$, respectively. Moreover, for any square matrix $A$, define its effective rank as $r_e(A) = \Tr(A)/\|A\|$. 
For any zero-mean random variables $X$ and $\tilde X$, we define the covariance matrix of $X$ as $\Sigma_X \triangleq \mathbb{E}[X X^\top]$, and the cross-covariance matrix of $X$ and $\tilde X$ as $\Sigma_{X, \tilde X} \triangleq \mathbb{E}[X \tilde X^\top]$.
Define $\hat \Sigma_{x,\tilde x}$ as $\hat\Sigma_{x,\tilde x} \triangleq 1/n \sum_{i \in [n]} x_i \tilde x_i^\top - 1/n/(n-1) \sum_{i \neq j} x_i \tilde x_j^\top$.

Here we prove results in \ref{sec: theory}. We consider minimizing the following linear loss function:
\begin{align*}
    \mL(G) \triangleq - \tr(G_1 \hat\Sigma_{x,\tilde x} G_2^\top) + \Pi(G),
\end{align*}
where $\Pi(G) = (\alpha/4) \|G G^\top - I_r\|_F^2$ with $\alpha > 0$. We also define the loss for mini-batch $\mB$ as $\mL(G; \mB) \triangleq - \tr(G_1 \hat\Sigma_{x,\tilde x,\mB} G_2^\top) + \Pi(G)$, where 
\begin{align*}
    \hat \Sigma_{x,\tilde x,\mB} \triangleq \frac{1}{|\mB|} \sum_{i \in \mB} x_i \tilde x_i^\top - \frac{1}{|\mB|(|\mB|-1)} \sum_{i \neq j; i, j \in \mB} x_i \tilde x_j^\top.
\end{align*}

\subsection{Differential Privacy of \dpclip}

Here we present the Gaussian mechanism, the theoretical foundation of DP-SGD and DP-Adam above.
\begin{definition}[Gaussian Mechanism]\citep{dwork_book}
Let $f:\mathcal{X}\rightarrow\mathbb{R}^d$ be an arbitrary d-dimensional function, i.e. $f(x)=[f_1(x), f_2(x), ..., f_d(x) ]$ for $x\in \mathcal{X}$. Then, the Gaussian mechanism with parameter $\sigma$ outputs,
\begin{align*}
    M(x)=[f_1(x) + Z_1, f_2(x)+Z_2, ..., f_d(x)+Z_d],
\end{align*}
where $Z_i\sim \mathcal{N}(0,\sigma^2)$ for $i\in [d].$
\end{definition}
\begin{definition}[$\ell_2$-sensitivity]
The $\ell_2$-sensitivity of a function $f:\mathcal{X}\rightarrow\mathcal{Y}$ is defined as,
\begin{align*}
    \Delta_2(f) = \max_{\text{adjacent } x_1, x_2\in \mathcal{X} } \|f(x_1)-f(x_2)\|_2.
\end{align*}
\end{definition}

We then present the privacy guarantee for \dpclip.
\begin{proposition}\label{prop: dp restatement}
     Choose for $b < n/10$. There exists universal constants $C_\epsilon, C_\sigma > 0$ such that
     for any $\epsilon \leq C_\epsilon b^2 T / n^2$ and $\delta > 0$, \dpclip is $(\epsilon,\delta)$-differentially private if we choose $\sigma \geq C_\sigma \sqrt{T \log(1/\delta)} / (n \epsilon)$.
\end{proposition}

\begin{proof}
    Note that at each iteration, we can view Algorithm \ref{alg:DP-GD} as a repeated composition of subsampling and Gaussian mechanism.
    Let $\mathcal{M} = \mathcal{M}_{T} \circ \mathcal{M}_{T-1} \circ \dots \circ \mathcal{M}_1$, where $\mathcal{M}_t \triangleq \mathcal{M}_{t,G} \circ \mathcal{M}_{t,s}$ is the composition of subsampling and Gaussian mechanism at $t$-th iteration.
    We first bound the $\ell_2$ sensitivity for Gaussian mechanism $\mathcal{M}_{t,G}$. Note that $\mathcal{M}_{t,G}$ depends on the mini-batch $\mB^{(t)} \subset [n]$. 
    We write $\bar g^{(t)} = \bar g^{(t)}((x_i, \tilde x_i)_{i \in \mB^{(t)}})$ to make explicit the dependence of $g^{(t)}$ on the pairs of data $(x_i, \tilde x_i)_{i \in \mB^{(t)}}$.
    Note that we can bound the $\ell_2$ sensitivity as
    \begin{align*}
        \max_{i \in \mB^{(t)}} \sup_{(x_i, \tilde x_i), (x_i', \tilde x_i') \in \R^{d_1+d_2}} |\bar g^{(t)}(\dots, (x_i, \tilde x_i), \dots) -\bar  g^{(t)}(\dots, (x_i', \tilde x_i'), \dots)| \leq 2c
    \end{align*}
    due to per-batch clipping.
    The rest of the proof follows from the result of privacy amplification by subsampling (Theorem 11 from \citet{bun2018composable}), and a similar argument in the proof of Lemma 3.1 from \citet{yang2022normalized}.
\end{proof}

\subsection{Optimization Error Bound}

In this subsection, we aim to derive the optimization error bound for $\dist(G^{(T)}, \hat G)$ given fixed pairs of data $\{(x_i, \tilde x_i)\}_{i=1}^n$, where the distance $\dist$ is defined as follows:
For any matrices $A \in \R^{r \times d}$ and $A' \in \R^{r \times d}$, define the distance as
\begin{align*}
    \dist(A, A') \triangleq \min_{O \in \mathbb{O}_{r,r}} \|O A - A'\|_F.
\end{align*}



\begin{assumption}[Local Directional Strong Convexity of $\mL$]\label{asm: hessian ap}
    Assume that there exists some $\gamma > 0$ such that for any $G$ satisfying $\|G - \hat G\|_F \leq \gamma$, the following inequalities hold for all $Z \in \R^{r\times d}$:
    \begin{align}
        \vect(Z)^\top \frac{\partial^2 \mL(G)}{\partial \vect(G) \partial \vect(G)^\top} \vect(Z) &\leq \beta_u \|Z\|_F^2.\label{eq: asm: beta upper ap}\\
        \vect(H_Z Z - \hat G)^\top \frac{\partial^2 \mL(G)}{\partial \vect(G) \partial \vect(G)^\top} \vect(H_Z Z - \hat G) &\geq \beta_l \|H Z - \hat G\|_F^2,\label{eq: asm: beta lower ap}
    \end{align}
    where $H_Z \triangleq \argmin_{O \in \mathbb{O}_{r,r}} \|O Z - \hat G\|_F$.
\end{assumption}

Before presenting the theorem and its proof, we list lemmas to be used in the proof. The proofs of lemmas are deferred to Section \ref{sec: proof of lemmas}.
\begin{lemma}\label{lem: one step linear convergence}
    Suppose that Assumption \ref{asm: hessian ap} holds with triple $(\beta_u, \beta_l, \gamma)$ and that
    \begin{align}
        \dist^2(G, \hat G) \leq \gamma^2.\label{eq: difference at t bounded by delta}
    \end{align}
    
    Let $H \triangleq \argmin_{H \in \mathbb{O}_{r,r}} \|H G - \hat G\|_F$. Define $\bar G \triangleq G - \eta \pG\mL(G)$ and $\tilde G \triangleq G - \eta g$, where $g \in \R^{r\times d}$ is any matrix.
    If $\eta \leq 1/\beta_u$, then,
    \begin{align*}
        \|H \bar G - \hat G\|_F^2 &\leq (1 - \eta \beta_l) \dist^2(G, \hat G),\\
        \|H \tilde G - \hat G\|_F^2 &\leq (1 - \eta \beta_l) \dist^2(G, \hat G) + 2 \eta \langle H \bar G - \hat G, \ \pG\mL(G) - g \rangle + \eta^2 \|g - \pG\mL(G)\|_F^2.
    \end{align*}
\end{lemma}

\begin{lemma}\label{lem: unbiased minibatch gradient}
    Let $\mB \subset [n]$ be a uniformly sampled random batch of size $b$ in $[n]$. Then,
    \begin{align*}
        \mathbb{E}_{\mB}[\pG\mL(G; \mB)] = \pG\mL(G)
    \end{align*}
    holds for all $G \in \R^{r \times d}$, where the expectation is taken with respect to subsampling.
\end{lemma}

Define $R \triangleq (\max_{i \in [n]} \|x_i\|) (\max_{i \in [n]} \|\tilde x_i\|)$.
\begin{lemma}\label{lem: gradient concentration}
    Fix $T > 0$.
    Suppose that $\max_{t \in [T]} \dist(G^{(t)}, \hat G)^2 \leq \gamma^2$ holds.
    Then,
    \begin{align*}
        \max_{t \in [T]} \|\pG\mL(G^{(t)}; \mB^{(t)}) - \pG\mL(G^{(t)})\|_F \lesssim (\|\hat G\|_F + \gamma) R \qty(\sqrt{\frac{(1 - b/n) \log(T(n+d))}{b}} + \frac{1}{b})
    \end{align*}
    holds with probability $1 - O((n+d)^{-1})$.
\end{lemma}

\begin{lemma}\label{lem: mini-batch gradient bound}
    Suppose that $x_i, \tilde x_i$ are generated according to the model in \eqref{model: dual spiked covariance model}.
    Suppose that $\max_{t \in [T]} \dist^2(G^{(t)}, \hat G) \leq \gamma^2$,
    where $\gamma$ satisfies $\gamma \leq 1 \wedge 1/\alpha$.
    Then,
    \begin{align*}
        \max_{t \in [T]} \|\pG\mL(G^{(t)}; \mB^{(t)})\|_F &\lesssim (\sqrt{r} \|\hat G\| + 1) R \sqrt{\frac{\log(T(n+d))}{b}} + \gamma \|\hat\Sigma_{x,\tilde x}\| + \alpha (\|\hat G\|^2 + 1) \gamma
    \end{align*}
    holds with probability $1 - O((n+d)^{-1})$.
\end{lemma}

Using above lemmas, we obtain the following theorem.
\begin{theorem}\label{thm: convergence to global minimum}
    Suppose that Assumption \ref{asm: hessian ap} holds with triple $(\beta_u, \beta_l, \gamma)$ and that
    \begin{align}
        \dist^2(G^{(0)}, \hat G) &\leq \frac{\gamma^2}{8}.\label{eq: initial value condition}
    \end{align}
    We obtain a sequence of representations $(G^{(t)})_{t\in[T]}$ from noisy mini-batch SGD according to Algorithm \ref{alg:DP-GD} with linear loss $\mL$.
    Set the clipping threshold $c$ and the mini-batch size $b$ as
    \begin{align}
        c &\gg (\sqrt{r} \|\hat G\| + 1) R \sqrt{\frac{\log(T(n+d))}{b}} + \gamma \|\hat\Sigma_{x,\tilde x}\| + \alpha (\|\hat G\|^2 + 1) \gamma,\label{eq: choice of c}\\
        b &\gg \frac{1}{\gamma^2} (\sqrt{r}\|\hat G\| + \gamma)^2 R^2 \log(T(n+d)).\label{eq: choice of b}
    \end{align}
    If $\eta > 0$ satisfies
    \begin{align}
        \eta \leq \min\qty{\frac{1}{2\beta_u}, \ \ \frac{\beta_l \gamma^2}{4\sigma^2 c^2 (2rd + 60\log(T(n+d))}},\label{eq: eta condition}
    \end{align}
    then,
    \begin{align*}
        \dist^2(G^{(T)}, \hat G) &\lesssim (1 - \eta \beta_l)^T \dist^2(G^{(0)}, \hat G) + \frac{\eta \sigma^2 c^2}{\beta_l} \qty(rd + \log(T(n+d)))\\
        &\quad+ \frac{\eta}{\beta_l} (\sqrt{r}\|\hat G\| + \gamma)^2 R^2 \qty(\sqrt{\frac{(1 - b/n) \log(T(n+d))}{b}} + \frac{1}{b})^2.
    \end{align*}
    holds with probability $1 - O((n+d)^{-1})$.
\end{theorem}

\begin{proof}[Proof of Theorem \ref{thm: convergence to global minimum}]\label{proof: thm: convergence to global minimum}
    For notational brevity, write $g^{(t)} = \pG\mL(G^{(t)}; \mB^{(t)})$.
    Define $H^{(t)} \triangleq \argmin_{H \in \mathbb{O}_{r,r}} \|H G^{(t)} - \hat G\|_F$. Also define $\tilde G^{(t+1)} := G^{(t)} - \eta g^{(t)}$.
    From \eqref{eq: choice of c} and Lemma \ref{lem: mini-batch gradient bound}, 
    \begin{align*}
        \max_{t \in [T]} \|\pG\mL(G^{(t)}; \mB^{(t)})\|_F \leq c
    \end{align*}
    holds with probability $1 - O((n+d)^{-1})$.
    Henceforth, we focus on this event, where $h^{(t)} = 1$ holds for all $t \in [T]$.
    Observe that
    \begin{align}
        \dist^2(G^{(t+1)}, \hat G) &\leq \|H^{(t)} G^{(t+1)} - \hat G\|_F^2\nonumber\\
        &= \|H^{(t)} \tilde G^{(t+1)} - \hat G - \eta \sigma c H^{(t)} \Gamma^{(t)}\|_F^2\nonumber\\
        &= \|H^{(t)} \tilde G^{(t+1)} - \hat G\|_F^2 -2 \eta \sigma c \tr( (H^{(t)} \tilde G^{(t+1)} - \hat G)^\top H^{(t)} \Gamma^{(t)}) + \eta^2 \sigma^2 c^2 \|\Gamma^{(t)}\|_F^2.\label{eq: dist t+1 decomposition}
    \end{align}
    Define $D^{(t+1)} \triangleq \|H^{(t)} \tilde G^{(t+1)} - \hat G\|_F$.
    Observe that
    \begin{align*}
        -2 \eta \sigma c \tr( (H^{(t)} \tilde G^{(t+1)} - \hat G)^\top H^{(t)} \Gamma^{(t)}) &= -2 \eta \sigma c \sum_{j \in [r], k \in [d_1]} (H^{(t) \top} (H^{(t)} \tilde G^{(t+1)} - \hat G))_{jk} (\Gamma^{(t)})_{jk}\\
        &\triangleq 2 \eta \sigma c D^{(t+1)} u^{(t)}.
    \end{align*}
    Also,
    \begin{align*}
        \eta^2 \sigma^2 c^2 \|\Gamma^{(t)}\|_F^2 &= \eta^2 \sigma^2 c^2 \tr(\Gamma^{(t) \top} \Gamma^{(t)}) = \eta^2 \sigma^2 c^2 \sum_{j \in [r], k \in [d]} (\Gamma^{(t)})_{ij}^2 \triangleq \eta^2 \sigma^2 c^2 v^{(t)}.
    \end{align*}
    Since $\vect(\Gamma^{(t)}) \sim N(0, I_{rd})$, $u^{(t)} \sim N(0, 1)$ and $v^{(t)} \sim \chi_{rd}^2$.
    For simplicity, write $d^{(t)} \triangleq \dist(G^{(t)}, \hat G)$.
    Now we have the following inequality:
    \begin{align*}
        d^{(t+1) 2} &\leq D^{(t+1) 2} + 2\eta \sigma c D^{(t+1)} u^{(t)} + \eta^2 \sigma^2 c^2 v^{(t)}.
    \end{align*}
    Let $V \triangleq \max_{t \in [T]} \|g^{(t)} - \pG\mL(G^{(t)})\|_F$. Using Lemma \ref{lem: one step linear convergence}, we obtain
    \begin{align}
        d^{(t+1) 2} &\leq (1 - \eta \beta_l) d^{(t) 2} + 2 \eta \langle H^{(t)} (G^{(t)} - \eta\pG\mL(G^{(t)})) - \hat G, \ \pG\mL(G^{(t)}) - g^{(t)} \rangle + \eta^2 V^2\nonumber\\
        &\quad+ 2\eta \sigma c D^{(t+1)} u^{(t)} + \eta^2 \sigma^2 c^2 v^{(t)},\label{eq: d and D}
    \end{align}
    which holds for all $t \in [T]$.
    
    We show by induction that the following inequality holds with probability $1 - O(s T^{-1} (n+d)^{-1})$ for any fixed $s \in [T]$:
    \begin{equation}
        d^{(s) 2} \leq {\min}\{\gamma^2, \ \ 4 (1 - \eta \beta_l)^s d^{(0) 2} + L\},\label{eq: induction target}
    \end{equation}
    where $L > 0$ is the solution of $L - C_2 \sqrt{L} - C_1^2 = 0$ with
    \begin{align*}
        C_1 &\triangleq \sqrt{\frac{\eta}{\beta_l} \sigma^2 c^2 (rd + 14 \log^2(T(n+d))) + 7 \frac{\eta}{\beta_l} V^2 \log(T(n+d))},\\
        C_2 &\triangleq 2 \sigma c \sqrt{2\log(T(n+d)) \frac{\eta}{\beta_l}} + 2 V \sqrt{\log(T(n+d)) \frac{\eta}{\beta_l}}.
    \end{align*}
    
    \paragraph{Step 1.} We start from $s=1$.
    From a standard concentration inequality for Gaussian random variables, (see, for example, Proposition 2.5 of \citet{wainwright2019high}.)
    \begin{align}
        u^{(0)} \leq \sqrt{2 \log(T(n+d))}\label{eq: u0 bound}
    \end{align}
    holds with probability at least $1 - T^{-1}(n+d)^{-1}$. 
    From a concentration bound for chi-squared distribution, (see, for example, Lemma 1 of \citet{laurent2000adaptive}.)
    \begin{align}
        v^{(0)} \leq rd + 2 \log(T(n+d))\label{eq: v0 bound}
    \end{align}
    holds with probability $1 - c T^{-1} (n+d)^{-1}$ for some universal constant $c > 0$.
    Note that Lemma \ref{lem: one step linear convergence} and Cauchy-Schwarz inequality yield
    \begin{align*}
        D^{(1) 2} &\leq (1 - \eta \beta_l) d^{(0) 2} + 2\eta d^{(0)} V + \eta^2 V^2\\
        &\leq (1 - \eta \beta_l) d^{(0) 2} + \eta \beta_l d^{(0) 2} + 2 \frac{\eta}{\beta_l} V^2\\
        &\leq 2 (1 - \eta \beta_l) d^{(0) 2} + 2 \frac{\eta}{\beta_l} V^2,
    \end{align*}
    where we used $2xy \leq x^2 + y^2$ in the second inequality and $\eta \beta_l \leq 1/2 \leq 1 - \eta \beta_l$ in the third inequality. Using $\sqrt{x + y} \leq \sqrt{x} + \sqrt{y}$ for $x, y \geq 0$, we further obtain
    \begin{align*}
        D^{(1)} &\leq \sqrt{2} (1 - \eta \beta_l)^{1/2} d^{(0)} + \sqrt{2 \frac{\eta}{\beta_l}} V.
    \end{align*}
    Combined with \eqref{eq: d and D}, \ref{eq: u0 bound} and \ref{eq: v0 bound}, we have
    \begin{align*}
        d^{(1) 2} &\leq D^{(1) 2} + 2\eta \sigma c D^{(1)} u^{(0)} + \eta^2 \sigma^2 c^2 v^{(0)}\\
        &\leq 2 (1 - \eta \beta_l) d^{(0) 2} + 2 \frac{\eta}{\beta_l} V^2 + 2\eta \sigma c \sqrt{2 \log(T(n+d))} D^{(1)} + \eta^2 \sigma^2 c^2 (rd + 2 \log(T(n+d)))\\
        &\leq 2 (1 - \eta \beta_l) d^{(0) 2} + 2 \frac{\eta}{\beta_l} V^2 + 4\eta \sigma c \sqrt{2 \log(T(n+d))} \qty((1 - \eta \beta_l)^{1/2} d^{(0)} + \sqrt{\frac{\eta}{\beta_l}} V)\\
        &\quad+ \eta^2 \sigma^2 c^2 (rd + 2 \log(T(n+d)))\\
        &\leq 4 (1 - \eta \beta_l) d^{(0) 2} + 4 \frac{\eta}{\beta_l} V^2 + \eta^2 \sigma^2 c^2 (rd + 10 \log(T(n+d))),
    \end{align*}
    where we used $2xy \leq x^2 + y^2$ in the last inequality. Notice that
    \begin{align*}
        \eta^2 \sigma^2 c^2 (rd + 10 \log(T(n+d))) + 4 \frac{\eta}{\beta_l} V^2 &\leq \frac{\eta \sigma^2 c^2}{\beta_l} (rd + 14 \log^2(T(n+d))) + 7 \frac{\eta}{\beta_l} V^2 \log(T(n+d))\\
        &= C_1^2 = L - C_2 \sqrt{L} \leq L.
    \end{align*}
    From \eqref{eq: initial value condition}, \eqref{eq: eta condition} and $L \leq \gamma^2 / 2$, which will be proved later,
    \begin{align*}
        4 (1 - \eta \beta_l) d^{(0) 2} + \eta^2 \sigma^2 c^2 (rd + 10 \log(T(n+d))) &\leq 4 \frac{\gamma^2}{8} + \frac{\gamma^2}{2} \leq \gamma^2.
    \end{align*}
    Therefore, we verify \eqref{eq: induction target} for $s=1$.    
        
    \paragraph{Step 2.} Fix $s \in [T]$. Suppose that \eqref{eq: induction target} holds for all $t$ satisfying $1 \leq t \leq s - 1$ on the event $E$. Examining the induction steps, we can show that the event $E$ occurs with probability at least $1 - (1 + c) (s-1)T^{-1} (n+d)^{-1}$.
    A similar concentration argument for $v^{(t)}$ gives,
    \begin{align}
        d^{(s) 2} &\leq (1 - \eta \beta_l) d^{(s-1) 2} + 2 \eta \langle H^{(s-1)} (G^{(s-1)} - \eta\pG\mL(G^{(s-1)})) - \hat G, \ \pG\mL(G^{(s-1)}) - g^{(s-1)} \rangle + \eta^2 V^2\nonumber\\
        &\quad+ 2\eta \sigma c D^{(s)} u^{(s-1)} + \eta^2 \sigma^2 c^2 (rd + 2 \log(T(n+d)))\label{eq: d and D for s}
    \end{align}
    holds with probability at least $1 - (1+c)(s-1)T^{-1} (n+d)^{-1} - c T^{-1} (n+d)^{-1}$.
    Applying \eqref{eq: d and D for s} repeatedly,
    \begin{align}
        d^{(s) 2} 
        &\leq (1 - \eta \beta_l)^s d^{(0) 2} + \frac{1}{\eta\beta_l} \qty{\eta^2 \sigma^2 c^2 (rd + 2 \log(T(n+d))) + \eta^2 V^2}\nonumber\\
        &\quad+ \underbrace{2 \eta \sigma c \sum_{t'=0}^{s-1} (1 - \eta \beta_l)^{s - t' - 1} D^{(t'+1)} u^{(t')}}_{T_1^{(s-1)}}\nonumber\\
        &\quad+ \underbrace{2 \eta \sum_{t'=0}^{s-1} (1 - \eta \beta_l)^{s - t' - 1} \langle H^{(t')} (G^{(t')} - \eta\pG\mL(G^{(t')})) - \hat G, \ \pG\mL(G^{(t')}) - g^{(t')} \rangle}_{T_2^{(s-1)}}.\label{eq: dist bound with u}
    \end{align}
    holds with probability at least $1 - (1+c)(s-1)T^{-1} (n+d)^{-1} - c T^{-1} (n+d)^{-1}$.

    Before bounding $T_1^{(s-1)}$ and $T_2^{(s-1)}$, we derive a concentration inequality for the following sum:
    \begin{align*}
         S^{(t-1)}_a \triangleq \sum_{t'=0}^{t-1} a^{t - t'-1} \langle H^{(t')} (G^{(t')} - \pG\mL(G^{(t')})) - \hat G, \ \pG\mL(G^{(t')}) - g^{(t')} \rangle,
    \end{align*}
    where $a \in (0, 1)$. 
    Fix $t > 0$. Let $\mathcal{F}^{(t')}$ be a filtration generated from $g^{(0)}, g^{(1)}, \dots, g^{(t')}$.
    Using Cauchy-Schwarz inequality and Lemma \ref{lem: one step linear convergence}, we obtain
    \begin{align*}
        |\langle H^{(t')} (G^{(t')} - \eta\pG\mL(G^{(t')})) - \hat G, \ \pG\mL(G^{(t')}) - g^{(t')} \rangle| \leq (1 - \eta \beta_l) d^{(t')} V.
    \end{align*}
    From Lemma \ref{lem: unbiased minibatch gradient}, $\mathbb{E}[g^{(t)} | \mathcal{F}^{(t-1)}] = \pG\mL(G^{(t)})$. Since $G^{(t')}$ and $\pG\mL(G^{(t')})$ are $\mathcal{F}^{(t'-1)}$-measurable,
    \begin{align*}
        \mathbb{E}[\langle H^{(t')} (G^{(t')} - \eta\pG\mL(G^{(t')})) - \hat G, \ \pG\mL(G^{(t')}) - g^{(t')} \rangle | \mathcal{F}^{(t'-1)}] &= 0.
    \end{align*}
    Thus $S^{(t-1)}_a$ is a sum of martingale difference sequence. Using Azuma-Hoeffding bound (See, for example, Corollary 2.20 of \citet{wainwright2019high}), we obtain
    \begin{align*}
        |S^{(t-1)}_a| &\leq \sqrt{\log(T(n+d)) \sum_{t'=0}^{t-1} a^{2t - 2t' - 2} (1 - \eta \beta_l) d^{(t') 2} V^2 }.
    \end{align*}
    with probability $1 - O(T^{-1}(n+d)^{-1})$.
    By a union bound argument, 
    \begin{align}
        \max_{t \in [T]} |S^{(t-1)}_a| &\leq \sqrt{\log(T(n+d)) \sum_{t'=0}^{t-1} a^{2t - 2t' - 2} (1 - \eta \beta_l) d^{(t') 2} V^2 }\label{eq: geometric series of inner product}
    \end{align}
    holds with probability $1 - O((n+d)^{-1})$.
    
    Here we bound the term $T_1^{(s-1)}$, since $u^{(0)}, u^{(1)}, u^{(2)}, \dots, u^{(s-1)}$ are i.i.d. standard normal random variables,
    \begin{align*}
        T_1^{(s-1)} &\leq 2 \eta \sigma c\sqrt{2 \Var\qty(\sum_{t'=0}^{s-1} (1 - \eta \beta_l)^{s-t'-1} D^{(t'+1)} u^{(t')}) \log(Td)}\nonumber\\
        &\leq 2 \eta \sigma c \sqrt{2 \log(Td) \sum_{t'=0}^{s-1} (1 - \eta \beta_l)^{2s - 2t' - 2} D^{(t'+1) 2}}.
    \end{align*}
    We bound $\sum_{t'=0}^{s-1} (1 - \eta \beta_l)^{2s - 2t' - 2} D^{(t'+1) 2}$. From Lemma \ref{lem: one step linear convergence} and \eqref{eq: induction target},
    \begin{align*}
        &\sum_{t'=0}^{s-1} (1 - \eta \beta_l)^{2s - 2t' - 2} D^{(t'+1) 2}\\
        &\quad\leq \sum_{t'=0}^{s-1} (1 - \eta \beta_l)^{2s - 2t' - 2} \qty( (1 - \eta \beta_l) d^{(t') 2} + 2\eta \langle H^{(t')} (G^{(t')} - \eta \pG\mL(G^{(t')})) - \hat G, \ \pG\mL(G^{(t')}) - \eta g^{(t')} \rangle + \eta^2 V^2)\\
        &\quad\leq \sum_{t'=0}^{s-1} 4 (1 - \eta \beta_l)^{2s - t' - 1} d^{(0) 2} + \sum_{t'=0}^{s-1} (1 - \eta \beta_l)^{2s - 2t' - 1} L + 2\eta S^{(s-1)}_{(1 - \eta \beta_l)^2} + \sum_{t'=0}^{s-1} (1 - \eta \beta_l)^{2s - 2t' - 2} \eta^2 V^2.
    \end{align*}
    Since $1 - \eta \beta_l \leq 1$,
    \begin{align}
        \sum_{t'=0}^{s-1} (1 - \eta \beta_l)^{4s-4t'-4} \leq \sum_{t'=0}^{s-1} (1 - \eta \beta_l)^{3s-3t'-3} \leq \sum_{t'=0}^{s-1} (1 - \eta \beta_l)^{2s-2t'-2} \leq \sum_{t'=0}^{s-1} (1 - \eta \beta_l)^{s-t'-1} \leq \frac{1}{\eta \beta_l}.\label{eq: geometric series}
    \end{align}
    Combined with Lemma \ref{lem: one step linear convergence}, \eqref{eq: induction target}, \eqref{eq: geometric series of inner product} and \eqref{eq: geometric series},
    \begin{align*}
        &\sum_{t'=0}^{s-1} (1 - \eta \beta_l)^{2s - 2t' - 2} D^{(t'+1) 2}\\
        &\quad\leq 4 (1 - \eta \beta_l)^s \frac{1}{\eta \beta_l} d^{(0) 2} + \frac{L}{\eta \beta_l} + 2\eta \sqrt{\log(T(n+d)) \sum_{t'=0}^{s-1} (1 - \eta \beta_l)^{4s - 4t' - 3} d^{(t') 2} V^2 } + \frac{\eta}{\beta_l} V^2\\
        &\quad\leq 4 (1 - \eta \beta_l)^s \frac{1}{\eta \beta_l} d^{(0) 2} + \frac{L}{\eta \beta_l} + \frac{\eta}{\beta_l} V^2 + 4 \eta V \sqrt{\log(T(n+d)) \sum_{t'=0}^{s-1} (1 - \eta \beta_l)^{4s - 3t' - 3} d^{(0) 2} }\\
        &\quad\quad+ 2\eta V \sqrt{\log(T(n+d)) \sum_{t'=0}^{s-1} (1 - \eta \beta_l)^{4s - 4t' - 3} L}\\
        &\quad= 4 (1 - \eta \beta_l)^s \frac{1}{\eta \beta_l} d^{(0) 2} + \frac{L}{\eta \beta_l} + \frac{\eta}{\beta_l} V^2 + 4 \eta V \sqrt{\log(T(n+d)) (1 - \eta \beta_l)^s d^{(0) 2} \sum_{t'=0}^{s-1} (1 - \eta \beta_l)^{3s - 3t' - 3} }\\
        &\quad\quad+ 2\eta V \sqrt{\log(T(n+d)) \sum_{t'=0}^{s-1} (1 - \eta \beta_l)^{4s - 4t' - 3} L}\\
        &\quad\leq 4 (1 - \eta \beta_l)^s \frac{1}{\eta \beta_l} d^{(0) 2} + \frac{L}{\eta \beta_l} + \frac{\eta}{\beta_l} V^2 + 4 \eta V \sqrt{\log(T(n+d)) (1 - \eta \beta_l)^s d^{(0) 2} \frac{1}{\eta \beta_l} }\\
        &\quad\quad+ 2\eta V \sqrt{\log(T(n+d)) L \frac{1}{\eta \beta_l}}\\
        &\quad\leq 8 (1 - \eta \beta_l)^s \frac{1}{\eta \beta_l} d^{(0) 2} + 2 \frac{L}{\eta \beta_l} + \frac{\eta}{\beta_l} V^2 + 3 \eta^2 V^2 \log(T(n+d))
    \end{align*}
    holds with probability at least $1 - T^{-1}(n+d)^{-1}$, where we used $\sqrt{x + y} \leq \sqrt{x} + \sqrt{y}$ and $2\sqrt{xy} \leq x + y$ for $x, y \geq 0$.
    Therefore,
    \begin{align}
        T_1^{(s-1)}
        &\leq 2 \eta \sigma c \sqrt{\log(T(n+d)) \qty(8 (1 - \eta \beta_l)^s \frac{1}{\eta \beta_l} d^{(0) 2} + 2 \frac{L}{\eta \beta_l} + \frac{\eta}{\beta_l} V^2 + 3 \eta^2 V^2 \log(T(n+d)))}\nonumber\\
        &\leq 4 \eta \sigma c \sqrt{2\log(T(n+d)) (1 - \eta \beta_l)^s \frac{1}{\eta \beta_l} d^{(0) 2}} + 2 \eta \sigma c \sqrt{2\log(T(n+d)) \frac{L}{\eta \beta_l}}\nonumber\\
        &\quad+ 2 \eta \sigma c \sqrt{\log(T(n+d)) \frac{\eta}{\beta_l} V^2} + 2 \eta \sigma c \sqrt{3 \log^2(T(n+d)) \eta^2 V^2}\nonumber\\
        &\leq (1 - \eta \beta_l)^s d^{(0) 2} + 8 \sigma^2 c^2 \frac{\eta}{\beta_l}\log(T(n+d)) + 2 \sigma c \sqrt{2\log(T(n+d)) \frac{\eta}{\beta_l} L}\nonumber\\
        &\quad+ \eta^2 \sigma^2 c^2 \log(T(n+d)) + \frac{\eta}{\beta_l} V^2 + 3 \eta^2 \sigma^2 c^2 \log^2(T(n+d)) + \eta^2 V^2,\label{eq: T1}
    \end{align}
    where we used $\sqrt{x+y} \leq \sqrt{x} + \sqrt{y}$ for $x, y \geq 0$ and $2 xy \leq x^2 + y^2$.
    
    We bound the term $T_2^{(s-1)}$. Using \eqref{eq: geometric series of inner product} and \eqref{eq: induction target},
    \begin{align}
        T_2^{(s-1)} &= 2 \eta S^{(s-1)}_{1 - \eta \beta_l}\nonumber\\
        &\leq 2\eta V \sqrt{\log(T(n+d)) \sum_{t'=0}^{s-1} (1 - \eta \beta_l)^{2s - 2t' - 1} d^{(t') 2} }\nonumber\\
        &\leq 4\eta V \sqrt{\log(T(n+d)) \sum_{t'=0}^{s-1} (1 - \eta \beta_l)^{2s - t' - 1} d^{(0) 2} } + 2\eta V \sqrt{\log(T(n+d)) \sum_{t'=0}^{s-1} (1 - \eta \beta_l)^{2s - 2t' - 1} L }\nonumber\\
        &\leq 4\eta V \sqrt{\log(T(n+d)) (1 - \eta \beta_l)^{s} \frac{1}{\eta \beta_l} d^{(0) 2} } + 2\eta V \sqrt{\log(T(n+d)) \frac{1}{\eta \beta_l} L }\nonumber\\
        &= 4 V \sqrt{\log(T(n+d)) (1 - \eta \beta_l)^{s} \frac{\eta}{\beta_l} d^{(0) 2} } + 2 V \sqrt{\log(T(n+d)) \frac{\eta}{\beta_l} L }\nonumber\\
        &\leq 4 \frac{\eta}{\beta_l} V^2 \log(T(n+d)) + 2 (1 - \eta \beta_l)^s d^{(0) 2} + 2 V \sqrt{\log(T(n+d)) \frac{\eta}{\beta_l} L },\label{eq: T2}
    \end{align}
    where we used $\sqrt{x + y} \leq \sqrt{x} + \sqrt{y}$ for $x, y \geq 0$, $2xy \leq x^2 + y^2$.
    From \eqref{eq: dist bound with u}, \eqref{eq: T1} and \eqref{eq: T2},
    \begin{align*}
        d^{(s) 2} &\leq (1 - \eta \beta_l)^s d^{(0) 2} + \frac{1}{\eta\beta_l} \qty{\eta^2 \sigma^2 c^2 (rd + 2 \log(T(n+d))) + \eta^2 V^2}\nonumber\\
        &\quad+ (1 - \eta \beta_l)^s d^{(0) 2} + 8 \sigma^2 c^2 \frac{\eta}{\beta_l}\log(T(n+d)) + 2 \sigma c \sqrt{2\log(T(n+d)) \frac{\eta}{\beta_l} L}\\
        &\quad+ \eta^2 \sigma^2 c^2 \log(T(n+d)) + \frac{\eta}{\beta_l} V^2 + 3 \eta^2 \sigma^2 c^2 \log^2(T(n+d)) + \eta^2 V^2\\
        &\quad+ 4 \frac{\eta}{\beta_l} V^2 \log(T(n+d)) + 2 (1 - \eta \beta_l)^s d^{(0) 2} + 2 V \sqrt{\log(T(n+d)) \frac{\eta}{\beta_l} L }\\
        &\leq 4(1 - \eta \beta_l)^s d^{(0) 2} + \frac{\eta}{\beta_l} \sigma^2 c^2 (rd + 14 \log^2(T(n+d))) + 7 \frac{\eta}{\beta_l} V^2 \log(T(n+d))\nonumber\\
        &\quad+ \qty(2 \sigma c \sqrt{2\log(T(n+d)) \frac{\eta}{\beta_l}} + 2 V \sqrt{\log(T(n+d)) \frac{\eta}{\beta_l}}) \sqrt{L}
    \end{align*}
    holds with probability at least $1 - (1+c)(s-1)T^{-1} (n+d)^{-1} - c T^{-1} (n+d)^{-1} - T^{-1} (n+d)^{-1} = 1 - (1+c) s T^{-1} (n+d)^{-1}$, where we used $2xy \leq x^2 + y^2$ in the third inequality.
    Note that
    \begin{align*}
        d^{(s) 2} &\leq 4 (1 - \eta \beta_l)^s d^{(0) 2} + L,
    \end{align*}
    since $C_2 \sqrt{L} + C_1^2 = L$. Combined with \eqref{eq: initial value condition} and $L \leq \gamma^2 / 2$, this further gives $d^{(s) 2} \leq \gamma^2$.
    
    Finally, we bound $L$.
    Solving $L = C_1^2 + C_2 \sqrt{L}$ gives
    \begin{align*}
        L &= \qty(\frac{C_2 + \sqrt{C_2^2 + 4 C_1^2}}{2})^2 \leq (C_1 + C_2)^2 \leq 2C_1^2 + 2C_2^2\\
        &= 2 \frac{\eta}{\beta_l} \sigma^2 c^2 (rd + 14 \log^2(T(n+d))) + 14 \frac{\eta}{\beta_l} V^2 \log(T(n+d))\\
        &\quad+ 2 \qty(2 \sigma c \sqrt{2\log(T(n+d)) \frac{\eta}{\beta_l}} + 2 V \sqrt{\log(T(n+d)) \frac{\eta}{\beta_l}})^2\\
        &\leq \frac{\eta \sigma^2 c^2}{\beta_l} \qty(2rd + 60 \log^2(T(n+d))) + 30 \frac{\eta}{\beta_l} V^2 \log(T(n+d)),
    \end{align*}
    where we used $\eta \beta_l \leq 1/2$. 
    Note that from \eqref{eq: choice of b} and Lemma \ref{lem: gradient concentration},
    \begin{align*}
        V = \max_{t \in [T]} \|g^{(t)} - \pG\mL(G^{(t)})\|_F \lesssim (\sqrt{r}\|\hat G\| + \gamma) L \sqrt{\frac{\log(T(n+d))}{b}} \ll \frac{\gamma}{\sqrt{\log(T(n+d))}}.
    \end{align*}
    Thus 
    \begin{align*}
        \frac{\eta}{\beta_l} V^2 \log(T(n+d)) \leq V^2 \log(T(n+d)) \leq \frac{\gamma^2}{4 \cdot 30}.
    \end{align*}
    From \eqref{eq: eta condition}, we can see that $L \leq \gamma^2/2$.
    Finally, since $4(1 - \eta\beta_l)^s d^{(0) 2} \leq 4 \gamma^2 / 8 = \gamma^2 /2$,
    \begin{align*}
        d^{(s) 2} \leq \min\qty{ \gamma^2, \ \ 4 (1 - \eta \beta_l)^s d^{(0) 2} + \frac{\eta \sigma^2 c^2}{\beta_l} \qty(2 rd + 60 \log^2(T(n+d))) + 30 \frac{\eta}{\beta_l} V^2 }
    \end{align*}
    holds with probability $1 - O(s T^{-1} (n+d)^{-1})$ for all $s \in [T]$. This concludes the induction.
    Again, Lemma \ref{lem: gradient concentration} concludes the proof.
\end{proof}

\subsection{Statistical Error Bound}

\begin{assumption}\label{asm: regime ap}
    Assume that $n \wedge d > r$ and
    \begin{align*}
        n \gg \qty(\alpha^2 + \frac{1}{\alpha^2}) r(r + s_1^{-2} r_e(\Sigma_\xi) + s_2^{-2} r_e(\Sigma_{\tilde \xi}))^2 \log^3(T(n+d)).
    \end{align*}
\end{assumption}

\begin{assumption}[Signal-to-noise Ratio]\label{asm: snr ap}
    Assume that $s_1^2 \wedge s_2^2 = \Omega(1)$.
\end{assumption}

\begin{assumption}[Signal Condition Number]\label{asm: signal condition number ap}
    Assume that $\kappa \triangleq \lambda_{\max}(\Sigma_z)/\lambda_{\min}(\Sigma_z) = O(1)$.
\end{assumption}

In this section, we let $G^* = [G_1^*, G_2^*]$ be the minimizer of the loss $\mathbb{E}[\mL(G)]$.
Also let $\hat G = [\hat G_1, \hat G_2]$ be the minimizer of the loss $\mL(G)$.
Before going into the proof of Theorem \ref{thm: convergence to global minimum}, we introduce lemmas to be used in the proof, which are based on Lemma B.7 in \citet{gao2021sparse}.
Write $\Sigma_{x,\tilde x} \triangleq \mathbb{E}[\hat\Sigma_{x,\tilde x}]$.

\begin{lemma}\label{lem: hessian}
    Suppose that Assumption \ref{asm: regime ap} holds.
    Choose $\gamma > 0$ such that
    \begin{align}
        \gamma \leq \min\qty{1, \ \ \frac{\lambda_r(\hat\Sigma_{x,\tilde x}) - \lambda_{r+1}(\hat\Sigma_{x,\tilde x})}{18\alpha (1 + \lambda_1(\hat\Sigma_{x,\tilde x})/\alpha)^{1/2}}}.\label{eq: choice of gamma}
    \end{align}
    Then, Assumption \ref{asm: hessian ap} holds with
    \begin{align*}
        \beta_u &\geq 8 \|\hat\Sigma_{x,\tilde x}\| + 12 \alpha, \ \ \beta_l \leq \frac{\lambda_r(\hat\Sigma_{x,\tilde x}) - \lambda_{r+1}(\hat\Sigma_{x,\tilde x})}{2}.
    \end{align*}
\end{lemma}

\begin{lemma}\label{lem: global minimum}
    Let $\mL'(G; \Sigma) := -\tr(G_1^\top \Sigma G_2) + (\alpha/4) \|G G^\top - I\|_F^2$.
    Suppose that $\lambda_r(\Sigma) > \lambda_{r+1}(\Sigma)$. Then, the minimizer $\hat G = [\hat G_1, \hat G_2]$ of $\mL'$ satisfies
    \begin{align*}
        \hat G_1 = \frac{1}{\sqrt{2}} V \qty(I_r + \frac{1}{\alpha} \Lambda_{[r]})^{1/2} P_{[r]}^\top, \ \ \hat G_2 = \frac{1}{\sqrt{2}} V \qty(I_r + \frac{1}{\alpha} \Lambda_{[r]})^{1/2} Q_{[r]}^\top,
    \end{align*}
    where $V \in \mathbb{O}_{r,r}$ is any orthogonal matrix, $\Lambda_{[r]}$ is the top-$r$ singular values of $\Sigma$, and $P_{[r]}$ and $Q_{[r]}$ are the corresponding left and right singular vectors, respectively.
\end{lemma}

\begin{theorem}\label{thm: convergence to population global minimum ap}
    Suppose that Assumptions \ref{asm: regime ap}, \ref{asm: snr ap} and \ref{asm: signal condition number ap} hold.
    Let $G_1^{(T)}$ and $G_2^{(T)}$ be the representation obtained from algorithm \ref{alg:DP-GD} under the loss $\mL(G)$.
    Suppose that initial representation $G^{(0)}$ satisfy
    \begin{align}
        \dist(G^{(0)}, \hat G) \ll \alpha \wedge \frac{1}{\alpha^2}.\label{eq: initial value condition theorem 2}
    \end{align}
    Choose $c \gg 1 + \alpha$ and $b = \ceil{\nu n}$, where $\nu \in (0, 1)$ is some constant.
    If $\eta > 0$ satisfies
    \begin{align}
        \eta \ll \min\qty{ 1 + \frac{1}{\alpha}, \ \ \frac{1}{\sigma^2 (\alpha^3 \vee \alpha^{-1/2}) (rd + \log(T(n+d))} },\label{eq: choice of eta}
    \end{align}
    then,
    \begin{align}
        &\min_{A \in \R^{r \times r}} \|A G_1^{(T)} - G_1^*\|_F \vee \min_{A \in \R^{r \times r}} \|A G_2^{(T)} - G_2^*\|_F\nonumber\\
        &\quad\lesssim \qty(1 - \frac{\eta}{4\kappa})^{T/2} \dist(G^{(0)}, \hat G) + \sigma (1 + \alpha) \sqrt{\eta (rd + \log(T(n+d)))}\nonumber\\
        &\quad\quad+ \qty(1 + \frac{1}{\alpha}) \sqrt{\frac{r(r + s_1^{-2} r_e(\Sigma_\xi) + s_2^{-2} r_e(\Sigma_{\tilde \xi}))^2 \log^3(n+d)}{n}}.\label{eq: final bound theorem 2}
    \end{align}
    holds with probability $1 - O((n+d)^{-1})$.
\end{theorem}

\begin{corollary}\label{cor: convergence to population global minimum ap}
    Assume the same conditions as in Theorem \ref{thm: convergence to population global minimum ap}.
    Choose $\eta$ as
    \begin{align*}
        \eta = \frac{1}{\sigma \sqrt{T(rd + \log(T(n+d)))}}.
    \end{align*}
    If $T$ satisfies
    \begin{align}
        T &\gg \frac{1}{\sigma^2 (1 + 1/\alpha)^2 (rd + \log(T(n+d)))}\\
        &\quad\vee \sigma^2 (\alpha^3 \vee \alpha^{-1/2})^2 (rd + \log(T(n+d))),\label{eq: T condition theorem 2}
    \end{align}
    then
    \begin{align}
        &\min_{A \in \R^{r \times r}} \|A G_1^{(T)} - G_1^*\|_F \vee \min_{A \in \R^{r \times r}} \|A G_2^{(T)} - G_2^*\|_F\nonumber\\
        &\quad\lesssim \exp(-\frac{\sqrt{T}}{8 \kappa \sigma \sqrt{rd + \log(T(n+d)))}}) \dist(G^{(0)}, \hat G) + (1 + \alpha) \sqrt{\frac{\sigma \sqrt{rd + \log(T(n+d))}}{\sqrt{T}}}\nonumber\\
        &\quad\quad+ \qty(\alpha + \frac{1}{\alpha}) \sqrt{\frac{r(r + s_1^{-2} r_e(\Sigma_\xi) + s_2^{-2} r_e(\Sigma_{\tilde \xi}))^2 \log^3(n+d)}{n}}\label{eq: cor B1}
    \end{align}
    holds with probability $1 - O((n+d)^{-1})$.
\end{corollary}

\begin{proof}[Proof of Corollary \ref{cor: convergence to population global minimum ap}]
    We directly use Theorem \ref{thm: convergence to population global minimum ap}.
    First, we see that condition \ref{eq: choice of eta} is satisfied from the condition \ref{eq: T condition theorem 2}.
    Note that
    \begin{align}
        \sigma (1 + \alpha) \sqrt{\eta (rd + \log(T(n+d)))} \lesssim (1 + \alpha) \sqrt{\frac{\sigma \sqrt{rd + \log(T(n+d))}}{\sqrt{T}}}.\label{eq: second term theorem 2}
    \end{align}
    The result follows from \eqref{eq: second term theorem 2} and \eqref{eq: final bound theorem 2}.
\end{proof}

\begin{corollary}[Restatement of Theorem \ref{thm: privacy utility tradeoff}]\label{cor: convergence to population global minimum 2 ap}
    Assume the same conditions as in Theorem \ref{thm: convergence to population global minimum ap} and Corollary \ref{cor: convergence to population global minimum ap}.
    Choose $\sigma = C_\sigma \sqrt{T \log(1/\delta)}/(n\epsilon)$ for some universal constant $C_\sigma$.
    If $T$ satisfies
    \begin{align*}
        T &\gg \qty(\frac{(n \epsilon)}{(1 + 1/\alpha) \sqrt{(rd + \log(T(n+d))) \log(1/\delta)}})^2\nonumber\\
        &\quad\vee \qty(\frac{(\alpha^3 \vee \alpha^{-1/2}) \sqrt{(rd + \log(T(n+d))) \log(1/\delta)}}{n \epsilon})^2,
    \end{align*}
    then
    \begin{align}
        &\min_{A \in \R^{r \times r}} \|A G_1^{(T)} - G_1^*\|_F \vee \min_{A \in \R^{r \times r}} \|A G_2^{(T)} - G_2^*\|_F\nonumber\\
        &\quad\lesssim \exp(-\frac{n\epsilon }{8 \kappa C_\sigma \sqrt{\log(1/\delta) \{rd + \log(T(n+d))}\}}) \dist(G^{(0)}, \hat G) + (1 + \alpha) \frac{(rd + \log(T(n+d)))^{1/4} \log^{1/4}(1/\delta)}{\sqrt{n \epsilon}}\nonumber\\
        &\quad\quad+ \qty(\alpha + \frac{1}{\alpha}) \sqrt{\frac{r(r + s_1^{-2} r_e(\Sigma_\xi) + s_2^{-2} r_e(\Sigma_{\tilde \xi}))^2 \log^3(n+d)}{n}}\label{eq: cor B2}
    \end{align}
    holds with probability $1 - O((n+d)^{-1})$.
\end{corollary}
Corollary \ref{cor: convergence to population global minimum 2 ap} directly follows from Corollary \ref{cor: convergence to population global minimum ap} with the choice $\sigma \gg \sqrt{T \log(1/\delta)}/(n \epsilon)$.

\begin{proof}[Proof of Theorem \ref{thm: convergence to population global minimum ap}]\label{proof: thm: convergence to population global minimum ap}
    From Lemma \ref{lem: global minimum}, we obtain
    \begin{align*}
        \hat G_1 &= \frac{1}{\sqrt{2}} \hat V \qty(I_r + \frac{1}{\alpha} \hat \Lambda_{[r]})^{1/2} \hat P_{[r]}^\top, \ \ \hat G_2 = \frac{1}{\sqrt{2}} \hat V \qty(I_r + \frac{1}{\alpha} \hat \Lambda_{[r]})^{1/2} \hat Q_{[r]}^\top,
    \end{align*}
    where $\hat V \in \mathbb{O}_{r,r}$ is any orthogonal matrix, $\hat\Lambda_{[r]} = \diag(\hat\lambda_1, \dots, \hat\lambda_r)$ is the top-$r$ singular values of $\hat\Sigma_{x, \tilde x}$, $\hat P_{[r]}$ and $\hat Q_{[r]}$ are the left and singular vectors of $\hat\Sigma_{x, \tilde x}$, respectively.
    Since $\mathbb{E}[\mL(G)] = \mL'(G; \Sigma_{x,\tilde x})$, we also obtain
    \begin{align*}
        G_1^* &= \frac{1}{\sqrt{2}} V \qty(I_r + \frac{1}{\alpha} \Lambda_{[r]})^{1/2} P_{[r]}^\top, \ \ G_2^* = \frac{1}{\sqrt{2}} V \qty(I_r + \frac{1}{\alpha} \Lambda_{[r]})^{1/2} Q_{[r]}^\top,
    \end{align*}
    where $V \in \mathbb{O}_{r,r}$ is any orthogonal matrix, $\Lambda_{[r]} = \diag(\lambda_1, \dots, \lambda_r)$ is the top-$r$ singular values of $\Sigma_{x, \tilde x}$, $P_{[r]}$ and $Q_{[r]}$ are the left and singular vectors of $\Sigma_{x, \tilde x}$, respectively.

    We first bound $\min_{A \in \R^{r\times r}} \|A \hat G_1 - G_1^*\|_F$.
    Let $H_P \triangleq \argmin_{O \in \mathbb{O}_{r,r}} \|O \hat P_{[r]}^\top - P_{[r]}^\top\|_F$.
    Using Theorem 3 from \citet{yu2015useful}, we have
    \begin{align}
        \|H_P \hat P_{[r]}^\top - P_{[r]}^\top\|_F &\lesssim \frac{(\lambda_1 + 1) }{\lambda_r^2 - \lambda_{r+1}^2} \sqrt{\frac{r(r + s_1^{-2} r_e(\Sigma_\xi) + s_2^{-2} r_e(\Sigma_{\tilde \xi})) \log(n+d)}{n}}\nonumber\\
        &\lesssim \sqrt{\frac{r(r + s_1^{-2} r_e(\Sigma_\xi) + s_2^{-2} r_e(\Sigma_{\tilde \xi})) \log(n+d)}{n}},\label{eq: distance between P hat and P}
    \end{align}
    where we used Assumption \ref{asm: signal condition number ap}, $\lambda_{r+1} = 0$ and $\lambda_1 = 1$.
    Let $A_P := V (I_r + (1/\alpha) \Lambda_{[r]})^{1/2} (I_r + (1/\alpha) \hat\Lambda_{[r]})^{-1/2} \hat V^{-1}$. Then, from Assumption \ref{asm: signal condition number ap},
    \begin{align}
        \|A_P\|^2 \leq \frac{1 + \lambda_1/\alpha}{1 + \hat\lambda_r/\alpha} \lesssim 1 \vee \kappa \lesssim 1.\label{eq: AP}
    \end{align}
    Moreover,
    \begin{align}
        \|A_P \hat G_1 - G_1^*\|_F &= \norm{V \qty(I_r + \frac{1}{\alpha} \Lambda_{[r]}) H_P \qty(I_r + \frac{1}{\alpha} \hat \Lambda_{[r]})^{-1/2} \hat V^\top \hat G_1 - G_1^*}_F\nonumber\\
        &= \norm{V \qty(I_r + \frac{1}{\alpha} \Lambda_{[r]}) (H_P \hat P_{[r]}^\top - P_{[r]}^\top)}_F\nonumber\\
        &\lesssim \qty(1 + \frac{1}{\alpha}) \sqrt{\frac{r(r + s_1^{-2} r_e(\Sigma_\xi) + s_2^{-2} r_e(\Sigma_{\tilde \xi})) \log(n+d)}{n}},\label{eq: distance between G hat and G star 2}
    \end{align}
    where the last inequality follows from \eqref{eq: distance between P hat and P}.

    Denote the $j$-th largest singular value of $\Sigma_{x,\tilde x}$ and $\hat\Sigma_{x,\tilde x}$ by $\lambda_j$ and $\hat\lambda_j$, respectively.
    Note that Lemma \ref{lem: cross-covariance concentration} and Assumption \ref{asm: regime ap} gives
    $\|\hat \Sigma_{x,\tilde x} - \Sigma_{x,\tilde x}\| \ll \|\Sigma_{x,\tilde x}\| = 1$ with probability $1 - O((n+d)^{-1})$. In particular, $\|\hat\Sigma_{x,\tilde x}\| \leq 2 \|\Sigma_{x,\tilde x}\| = 2$.
    Furthermore, from Weyl's inequality, we also have $\max_{j \in [d]} |\hat \lambda_j - \lambda_j| \ll (\lambda_1/\lambda_j) \lambda_j \leq \kappa \lambda_j$.
    Thus, Assumption \ref{asm: signal condition number ap} gives
    \begin{align*}
        \hat \lambda_r - \hat \lambda_{r+1} \geq \lambda_r - \lambda_{r+1} - |\hat\lambda_r - \lambda_r| - |\hat \lambda_{r+1} - \lambda_{r+1}| \geq \frac{\lambda_r - \lambda_{r+1}}{2} = \frac{1}{2\kappa}.
    \end{align*}
    Choose $\gamma > 0$ such that
    \begin{align*}
        \gamma = \frac{1}{36 \kappa (\alpha\vee 1) (1 + 1/(2\alpha))^{1/2}}.
    \end{align*}
    Then, $\gamma$ satisfies the condition of Lemma \ref{lem: hessian} with probability $1 - O((n+d)^{-1})$.
    Thus, on this event, Assumption \ref{asm: hessian ap} holds for $\mL(G)$ with
    \begin{align}
        \beta_u \geq 8 \hat \lambda_1 + 12 \alpha, \ \ \beta_l \leq \frac{\hat\lambda_r - \hat\lambda_{r+1}}{2}.\label{eq: beta u beta l}
    \end{align}
    Choose $\beta_u = 16 + 12 \alpha$ and $\beta_l = (\lambda_r - \lambda_{r+1}) / 4 = 1 / (4\kappa)$, which satisfies \eqref{eq: beta u beta l} from the above arguments.

    From Lemma \ref{lem: good event},
    \begin{align*}
        R &\lesssim \sqrt{r + s_1^{-2} r_e(\Sigma_\xi)} \sqrt{r + s_2^{-2} r_e(\Sigma_{\tilde \xi})} \log(n+d)\\
        &\lesssim (r + s_1^{-2} r_e(\Sigma_\xi) + s_2^{-2} r_e(\Sigma_{\tilde \xi})) \log(n+d)
    \end{align*}
    holds with probability $1 - O((n+d)^{-1})$. 
    Choose $b = \ceil{\nu n}$.
    Since $\|\hat G\|^2 \leq 1 + \|\hat\Sigma_{x,\tilde x}\|/\alpha \leq 1 + 2/\alpha$,
    \begin{align}
        &(\sqrt{r} \|\hat G\| + 1) R \sqrt{\frac{\log(T(n+d))}{b}} + \gamma \|\hat\Sigma_{x,\tilde x}\| + \alpha (\|\hat G\|^2 + 1) \gamma\nonumber\\
        &\quad\lesssim \sqrt{r \qty(1 + \frac{1}{\alpha})} (r + s_1^{-2} r_e(\Sigma_\xi) + s_2^{-2} r_e(\Sigma_{\tilde \xi})) \log(n+d) \sqrt{\frac{\log(T(n+d))}{n}}  + 1 + \alpha \qty(1 + \frac{1}{\alpha})\nonumber\\
        &\quad\lesssim \sqrt{\qty(1 + \frac{1}{\alpha})} \frac{\sqrt{r} (r + s_1^{-2} r_e(\Sigma_\xi) + s_2^{-2} r_e(\Sigma_{\tilde \xi})) \log^{3/2}(n+d)}{\sqrt{n}} + 1 + \alpha\nonumber\\
        &\quad\lesssim 1 + \alpha,\label{eq: c lower bound}
    \end{align}
    where the last inequality follows from Assumption \ref{asm: regime ap}.
    Also note that
    \begin{align}
        &\frac{1}{\gamma^2} (\sqrt{r}\|\hat G\| + \gamma)^2 R^2 \log(T(n+d))\nonumber\\
        &\quad\lesssim \kappa^2 (\alpha^2 \vee 1) \qty(1 + \frac{1}{\alpha}) r\qty(1 + \frac{1}{\alpha}) (r + s_1^{-2} r_e(\Sigma_\xi) + s_2^{-2} r_e(\Sigma_{\tilde \xi}))^2 \log^2(n+d) \log(T(n+d))\nonumber\\
        &\quad\lesssim \qty(\alpha^2 + \frac{1}{\alpha^2}) r (r + s_1^{-2} r_e(\Sigma_\xi) + s_2^{-2} r_e(\Sigma_{\tilde \xi}))^2 \log^3(T(n+d))\nonumber\\
        &\quad\ll n,\label{eq: b lower bound}
    \end{align}
    where the last inequality follows again from Assumption \ref{asm: regime ap}.
    Choose $c \gg 1 + \alpha$.
    From \eqref{eq: c lower bound}, \eqref{eq: b lower bound} and $b = \ceil{\nu n}$, we verify that \eqref{eq: choice of c} and \eqref{eq: choice of b} are satisfied.

    From Theorem \ref{thm: convergence to global minimum}, if 
    \begin{align*}
        \eta \leq \min\qty{ \frac{1}{2 (16 + 12 \alpha)}, \ \ \frac{\lambda_r \gamma^2}{16\sigma^2 c^2 (2 rd + 60 \log(T(n+d))} },
    \end{align*}
    then the following bound holds with probability $1 - O((n+d)^{-1})$:
    \begin{align*}
        \dist^2(G^{(T)}, \hat G) &\lesssim (1 - \eta \beta_l)^T \dist^2(G^{(0)}, \hat G) + \frac{\eta \sigma^2 c^2}{\beta_l} \qty(rd + \log(T(n+d)))\\
        &\quad+ \frac{\eta}{\beta_l} (\sqrt{r}\|\hat G\| + \gamma)^2 R^2 \qty(\sqrt{\frac{(1 - b/n) \log(T(n+d))}{b}} + \frac{1}{b})^2.
    \end{align*}
    Substituting the values of $c$ and $b$ with a similar argument as in \eqref{eq: c lower bound} combined with  $\sqrt{x + y} \leq \sqrt{x} + \sqrt{y}$ gives
    \begin{align}
        \dist(G^{(T)}, \hat G) &\lesssim (1 - \eta \beta_l)^{T/2} \dist(G^{(0)}, \hat G) + \sigma c \sqrt{\frac{\eta}{\beta_l} (rd + \log(T(n+d)))}\nonumber\\
        &\quad+ \sqrt{\frac{\eta}{\beta_l} r\qty(1 + \frac{1}{\alpha})} (r + r_e(\Sigma_\xi) + r_e(\Sigma_{\tilde \xi})) \log(n+d) \sqrt{\frac{\log(T(n+d))}{n}}\nonumber\\
        &\lesssim (1 - \eta \beta_l)^{T/2} \dist(G^{(0)}, \hat G) + (1 + \alpha) \sigma \sqrt{\frac{\eta}{\beta_l} (rd + \log(T(n+d)))}\\
        &\quad+ \sqrt{\frac{\eta}{\beta_l} r\qty(1 + \frac{1}{\alpha})} \frac{(r + r_e(\Sigma_\xi) + r_e(\Sigma_{\tilde \xi})) \log^{3/2}(T(n+d))}{\sqrt{n}}.\label{eq: dist G T and hat G}
    \end{align}    
    
    Finally, note that
    \begin{align*}
        \min_{A \in \R^{r \times r}} \|A G_1^{(T)} - G_1^*\|_F &\leq \min_{A \in \R^{r \times r}} \|A G_1^{(T)} - A_P \hat G_1\|_F + \|A_P \hat G_1 - G_1^*\|_F\\
        &= \min_{A \in \R^{r \times r}} \|A_P (A_P^{-1} A G_1^{(T)} - \hat G_1)\|_F + \|A_P \hat G_1 - G_1^*\|_F\\
        &\leq \|A_P\| \min_{O \in \mathbb{O}_{r,r}} \|O G_1^{(T)} - \hat G_1\|_F + \|A_P \hat G_1 - G_1^*\|_F\\
        &\lesssim \dist(G^{(T)}, \hat G) + \|A_P \hat G_1 - G_1^*\|_F,
    \end{align*}
    where the last inequality follows from \eqref{eq: AP}.
    Using \eqref{eq: distance between G hat and G star 2} and \eqref{eq: dist G T and hat G}, we obtain
    \begin{align*}
        \min_{A \in \R^{r \times r}} \|A G_1^{(T)} - G_1^*\|_F &\leq \min_{A \in \R^{r \times r}} \|A G_1^{(T)} - A_P \hat G_1\|_F + \|A_P \hat G_1 - G_1^*\|_F\\
        &\lesssim (1 - \eta \beta_l)^{T/2} \dist(G^{(0)}, \hat G) + \sigma (1 + \alpha) \sqrt{\frac{\eta}{\beta_l} (rd + \log(T(n+d)))}\\
        &\quad+ \sqrt{\frac{\eta}{\beta_l} r\qty(1 + \frac{1}{\alpha})} \frac{(r + r_e(\Sigma_\xi) + r_e(\Sigma_{\tilde \xi})) \log^{3/2}(T(n+d))}{\sqrt{n}}\\
        &\quad+ \qty(1 + \frac{1}{\alpha}) \sqrt{\frac{r(r + s_1^{-2} r_e(\Sigma_\xi) + s_2^{-2} r_e(\Sigma_{\tilde \xi})) \log(n+d)}{n}}\\
        &\lesssim (1 - \eta \beta_l)^{T/2} \dist(G^{(0)}, \hat G) + \sigma (1 + \alpha) \sqrt{\frac{\eta}{\beta_l} (rd + \log(T(n+d)))}\\
        &\quad+ \qty(1 + \frac{1}{\alpha}) \sqrt{\frac{r(r + s_1^{-2} r_e(\Sigma_\xi) + s_2^{-2} r_e(\Sigma_{\tilde \xi}))^2 \log^3(n+d)}{n}}.
    \end{align*}

    A symmetric argument for $\hat G_2$ and $G_2^*$ gives the desired result.
\end{proof}

\section{Proof of Lemmas}\label{sec: proof of lemmas}

\begin{proof}[Proof of Lemma \ref{lem: one step linear convergence}]
    We first show the following inequality, as in the proof of Lemma 4 in \citet{chi2019nonconvex}.
    \begin{align}
        2 \left\langle H \pG\mL(G), H G - \hat G \right\rangle \geq \frac{1}{\beta_u} \|\pG\mL(G)\|_F^2 + \beta_l \|H G - \hat G\|_F^2.\label{eq: G t+1 - hat G inner product lower bound}
    \end{align}
    Note that $H \pG\mL(G) = \pG\mL(H G)$.
    Applying Taylor series expansion to $\mL(\hat G)$, we obtain
    \begin{align*}
        &\mL(\hat G) = \mL(H G) - \left\langle H \pG\mL(G), H G - \hat G \right\rangle + \frac{1}{2} \vect(H G - \hat G)^\top \frac{\partial^2 \mL(\check G)}{\partial \vect(G) \partial \vect(G)^\top} \vect(H G - \hat G),
    \end{align*}
    where $\check G \triangleq H G + \tau(\hat G - H G)$ with some $\tau \in [0, 1]$.
    We can see that
    \begin{align*}
        \|\check G - \hat G\|_F^2 = (1 - \tau) \|H G - \hat G\|_F^2 \leq \gamma^2.
    \end{align*}
    From \eqref{eq: asm: beta lower ap},
    \begin{align}
        \mL(\hat G) &\geq \mL(H G) - \left\langle H \pG\mL(G), H G - \hat G \right\rangle + \frac{\beta_l}{2} \|H G - \hat G\|_F^2.\label{eq: loss expansion lower}
    \end{align}
    Furthermore, from \eqref{eq: asm: beta upper ap} and \eqref{eq: difference at t bounded by delta},
    \begin{align}
        \mL(\hat G) - \mL(H G) &\leq \mL\qty(H G - \frac{1}{\beta_u} \pG\mL(H G)) - \mL(H G)\nonumber\\
        &\leq - \frac{1}{\beta_u} \langle \pG\mL(HG), \pG\mL(HG) \rangle + \frac{\beta_u}{2} \norm{\frac{1}{\beta_u} \pG\mL(HG)}_F^2\nonumber\\
        &= -\frac{1}{2\beta_u} \|\pG\mL(HG)\|_F^2 = -\frac{1}{2\beta_u} \|\pG\mL(G)\|_F^2,\label{eq: loss expansion upper}
    \end{align}
    where the second inequality follows from the characterization of smoothness (Theorem 5.8 of \citet{beck2017first}.)
    From \eqref{eq: loss expansion lower} and \eqref{eq: loss expansion upper}, we obtain
    \begin{align*}
        - \left\langle H \pG\mL(G), H G - \hat G \right\rangle + \frac{\beta_l}{2} \|H G - \hat G\|_F^2 &\leq \mL(\hat G) - \mL(H G)\\
        &\leq -\frac{1}{2\beta_u} \norm{\pG\mL(G)}_F^2.
    \end{align*}
    This proves \eqref{eq: G t+1 - hat G inner product lower bound}.
    
    Next, we bound $\|H \bar G - \hat G\|_F$. Observe that
    \begin{align}
        \|H \bar G - \hat G\|_F^2 &= \|H G - \hat G - \eta H \pG\mL(G)\|_F^2\nonumber\\
        &= \dist^2(G, \hat G) + \eta^2 \|\pG\mL(G)\|_F^2 - 2 \eta \left\langle \pG\mL(G), H G - \hat G \right\rangle\nonumber\\
        &\leq (1 - \eta \beta_l) \dist^2(G, \hat G) + \eta \qty(\eta - \frac{1}{\beta_u}) \|\pG\mL(G)\|_F^2\nonumber\\
        &\leq (1 - \eta \beta_l) \dist^2(G, \hat G),\label{eq: T1 bound}
    \end{align}
    where the last inequality follows from $\eta \leq 1/\beta_u$. 
    
    Using \eqref{eq: T1 bound}, we further obtain
    \begin{align*}
        \|H \tilde G - \hat G\|_F^2 &= \|H \bar G - \hat G + \eta H (\pG\mL(G) - g)\|_F^2\\
        &= \|H \bar G - \hat G\|_F^2 + 2 \eta \langle H \bar G - \hat G, \ \pG\mL(G) - g \rangle + \eta^2 \|g - \pG\mL(G)\|_F^2\\
        &\leq (1 - \eta \beta_l) \dist^2(G, \hat G) + 2 \eta \langle H \bar G - \hat G, \ \pG\mL(G) - g \rangle + \eta^2 \|g - \pG\mL(G)\|_F^2.
    \end{align*}
    This concludes the proof.
\end{proof}

\begin{proof}[Proof of Lemma \ref{lem: unbiased minibatch gradient}]
    Observe that
    \begin{align*}
        \mathbb{E}_{\mB}\qty[\frac{1}{b}\sum_{i\in \mB} x_i \tilde x_i^\top ]
        &= \frac{1}{b} \frac{1}{\binom{n}{b}} \sum_{\substack{\mB' \subset [n]\\|\mB'| = b}} \sum_{i \in \mB'} x_i \tilde x_i^\top\\
        &= \frac{1}{b} \frac{1}{\binom{n}{b}} \sum_{i \in [n]} x_i \tilde x_i^\top \sum_{\substack{\mB' \subset [n]\\|\mB'| = b}} \1\{i \in \mB'\}\\
        &= \frac{1}{b} \frac{\binom{n-1}{b-1}}{\binom{n}{b}} \sum_{i \in [n]} x_i \tilde x_i^\top\\
        &= \frac{1}{n} \sum_{i \in [n]} x_i \tilde x_i^\top.
    \end{align*}
    Similarly,
    \begin{align*}
        \mathbb{E}_{\mB}\qty[\frac{1}{b(b-1)}\sum_{\substack{i, j \in \mB\\i \neq j}} x_i \tilde x_j^\top ]
        &= \frac{1}{b(b-1)} \frac{1}{\binom{n}{b}} \sum_{\substack{\mB' \subset [n]\\|\mB'| = b}} \sum_{\substack{i,j \in \mB'\\i \neq j}} x_i \tilde x_j^\top\\
        &= \frac{1}{b(b-1)} \frac{\binom{n-2}{b-2}}{\binom{n}{b}} \sum_{\substack{i,j \in [n]\\i \neq j}} x_i \tilde x_j^\top\\
        &= \frac{1}{n(n-1)} \sum_{\substack{i,j \in [n]\\i \neq j}} x_i \tilde x_j^\top.
    \end{align*}
    Thus $\mathbb{E}_\mB[\hat \Sigma_{x,\tilde x,\mB}] = \mathbb{E}_\mB[\hat\Sigma_{x,\tilde x}]$ and hence
    \begin{align}
        \mathbb{E}_{\mB} [\mL(G; \mB)] = \mL(G).\label{eq: unbiased}
    \end{align}
    Taking derivative with $G$ in \eqref{eq: unbiased} concludes the proof.
\end{proof}

\begin{proof}[Proof of Lemma \ref{lem: gradient concentration}]
    Note that $\pG\mL(G^{(t)}; \mB^{(t)}) - \pG\mL(G^{(t)}; \mB^{(t)}) = \pG (-{\tr}(G_1^{(t)} \hat \Sigma_{x,\tilde x} G_2^{(t)}) + {\tr}(G_1^{(t)} \hat \Sigma_{x,\tilde x, \mB^{(t)}} G_2^{(t)}))$.
    Thus
    \begin{align*}
        &\|\pG\mL(G^{(t)}; \mB^{(t)}) - \pG\mL(G^{(t)}; \mB^{(t)})\|_F\\
        &\quad= \|G_1^{(t)} (\hat \Sigma_{x,\tilde x} - \hat \Sigma_{x,\tilde x,\mB^{(t)}})\|_F + \|G_2^{(t)} (\hat \Sigma_{x,\tilde x} - \hat \Sigma_{x,\tilde x,\mB^{(t)}})^\top\|_F\\
        &\quad\leq (\|G_1^{(t)}\|_F + \|G_2^{(t)}\|_F) \|\hat \Sigma_{x,\tilde x} - \hat \Sigma_{x,\tilde x,\mB^{(t)}}\|\\
        &\quad\leq (\|\hat G\|_F + \dist(G^{(t)}, \hat G)) \|\hat \Sigma_{x,\tilde x} - \hat \Sigma_{x,\tilde x,\mB^{(t)}}\|\\
        &\quad\leq (\|\hat G\|_F + \gamma) \|\hat \Sigma_{x,\tilde x} - \hat \Sigma_{x,\tilde x,\mB^{(t)}}\|.
    \end{align*}
    
    We first bound $\|(1/b) \sum_{i \in \mB^{(t)}} x_i \tilde x_i^\top - (1/n) \sum_{i \in [n]} x_i \tilde x_i^\top\|$.
    Write $x_i = (x_{i1}, \dots, x_{id_1})$ and $\tilde x_i = (\tilde x_{i1}, \dots, \tilde x_{id_2})$. For any fixed $k \in [d_1]$ and $\ell \in [d_2]$, using Lemma \ref{lem: subsampling concentration}, we have
    \begin{align*}
        \abs{\frac{1}{b} \sum_{i \in \mB^{(t)}} x_{ik} \tilde x_{il} - \frac{1}{n} \sum_{i \in [n]} x_{ik} \tilde x_{il}} &\leq C \max_{i \in [n]} |x_{ik} \tilde x_{il}| \sqrt{\frac{(1 - b/n) \log(Trd(n + d))}{b}}\\
        &\leq C R \sqrt{\frac{(1 - b/n) \log(T(n+d))}{b}}
    \end{align*}
    with probability $1 - O(T^{-1} (rd)^{-1} (n+d)^{-1})$, where $C > 0$ is a universal constant.
    Note that operator norm of a matrix is bounded by the maximum element of the matrix. By a union bound argument,
    \begin{align*}
        \norm{\frac{1}{b} \sum_{i \in \mB^{(t)}} x_i \tilde x_i^\top - \frac{1}{n} \sum_{i \in [n]} x_i \tilde x_i^\top} &\leq \max_{k \in [d_1], l \in [d_2]} \abs{\frac{1}{b} \sum_{i \in \mB^{(t)}} x_{ik} \tilde x_{il} - \frac{1}{n} \sum_{i \in [n]} x_{ik} \tilde x_{il}}\\
        &\leq C R \sqrt{\frac{(1 - b/n) \log(Trd(n + d))}{b}}
    \end{align*}
    holds with probability $1 - O(T^{-1} (n+d)^{-1})$.

    Let $ \mu_{\mB^{(t)}} \triangleq (1/b) \sum_{i \in \mB^{(t)}} x_i$ and $\tilde  \mu_{\mB^{(t)}} \triangleq (1/b) \sum_{i \in \mB^{(t)}} \tilde x_i$.
    Also let $\mu \triangleq (1/n) \sum_{i \in [n]} x_i$ and $\tilde \mu \triangleq (1/n) \sum_{i \in [n]} \tilde x_i$.
    Next we bound $\| \mu_{\mB^{(t)}} \tilde  \mu_{\mB^{(t)}}^\top - \mu \tilde \mu^\top\|$.

    Again from Lemma \ref{lem: subsampling concentration}, for any fixed $j \in [d_1]$,
    \begin{align*}
        |e_j^\top ( \mu_{\mB^{(t)}} - \mu)| \leq C' \max_{i \in [n]} |e_j^\top x_i| \sqrt{\frac{(1 - b/n) \log(T d_1(n+d))}{b}}
    \end{align*}
    holds with probability $1 - O(T^{-1} d_1^{-1} (n+d)^{-1})$, where $C' > 0$ is some universal constant.
    By a union bound argument, we obtain
    \begin{align*}
        \norm{ \mu_{\mB^{(t)}} - \mu} \leq C' \max_{i \in [n]} \|x_i\| \sqrt{\frac{(1 - b/n) \log(T(n+d))}{b}}
    \end{align*}
    holds with probability $1 - O(T^{-1} (n+d)^{-1})$. Similarly,
    \begin{align*}
        \norm{\tilde  \mu_{\mB^{(t)}} - \tilde \mu} \leq C' \max_{i \in [n]} \|\tilde x_i\| \sqrt{\frac{(1 - b/n) \log(n+d)}{b}}
    \end{align*}
    holds with probability $1 - O((n+d)^{-1})$.
    Thus,
    \begin{align*}
        \| \mu_{\mB^{(t)}} \tilde  \mu_{\mB^{(t)}}^\top - \mu \tilde \mu^\top\| &\leq \|\mu -  \mu_{\mB^{(t)}}\| \|\tilde  \mu_{\mB^{(t)}}\| + \|\mu\| \|\tilde \mu - \tilde  \mu_{\mB^{(t)}}\|\\
        &\leq 2 C' (\max_{i \in [n]} \|x_i\|) (\max_{i \in [n]} \|\tilde x_i\|) \sqrt{\frac{(1 - b/n) \log(T(n+d))}{b}},
    \end{align*}
    where we used $\|\tilde  \mu_{\mB^{(t)}}\| \leq \max_{i \in [n]} \|\tilde x_i\|$ and $\|\mu\| \leq \max_{i \in [n]} \|x_i\|$. Therefore,
    \begin{align*}
        \| \hat \Sigma_{x,\tilde x,\mB^{(t)}} - \hat \Sigma_{x,\tilde x} \| &= \norm{\frac{1}{b-1} \sum_{i \in \mB^{(t)}} x_i \tilde x_i^\top - \frac{1}{n-1} \sum_{i \in [n]} x_i \tilde x_i^\top - \frac{b}{b-1}  \mu_{\mB^{(t)}} \tilde  \mu_{\mB^{(t)}}^\top + \frac{n}{n-1} \mu \tilde \mu^\top}\\
        &\leq \norm{\frac{1}{b} \sum_{i \in \mB^{(t)}} x_i \tilde x_i^\top - \frac{1}{n} \sum_{i \in [n]} x_i \tilde x_i^\top} + \| \mu_{\mB^{(t)}} \tilde  \mu_{\mB^{(t)}}^\top - \mu \tilde \mu^\top\|\\
        &\quad+ \norm{\frac{1}{b} \sum_{i \in \mB^{(t)}} x_i \tilde x_i^\top - \frac{1}{b-1} \sum_{i \in \mB^{(t)}} x_i \tilde x_i^\top} + \norm{\frac{1}{n} \sum_{i \in [n]} x_i \tilde x_i^\top + \frac{1}{n-1} \sum_{i \in [n]} x_i \tilde x_i^\top}\\
        &\leq (C + 2C') R \sqrt{\frac{(1 - b/n) \log(T(n+d))}{b}}\\
        &\quad+ \frac{1}{b-1} \max_{i \in \mB^{(t)}} \|x_i\| \|\tilde x_i\| + \frac{1}{n-1} \max_{i \in [n]} \|x_i\| \|\tilde x_i\|\\
        &\leq (C + 2C' + 4) R  \qty(\sqrt{\frac{(1 - b/n) \log(T(n+d))}{b}} + \frac{1}{b})
    \end{align*}
    holds with probability $1 - O(T^{-1} (n+d)^{-1})$.
    A union bound argument for $t \in [T]$ concludes the proof.
\end{proof}

\begin{proof}[Proof of Lemma \ref{lem: mini-batch gradient bound}]
    Note that
    \begin{align*}
        \max_{t \in [T]} \|\pG\mL(G^{(t)}; \mB)\|_F &= \max_{t \in [T]} \|\pG\mL(G^{(t)}; \mB) - \pG\mL(\hat G)\|_F\\
        &\leq \max_{t \in [T]} \underbrace{\|\pG\mL(G^{(t)}; \mB) - \pG\mL(G^{(t)})\|_F}_{=: T_1^{(t)}} + \max_{t \in [T]} \underbrace{\|\pG\mL(G^{(t)}) - \pG\mL(\hat G)\|_F}_{=: T_2^{(t)}}.
    \end{align*}
    We can bound the term $T_1^{(t)}$ by Lemma \ref{lem: gradient concentration} as
    \begin{align}
        \max_{t \in [T]} T_1^{(t)} \lesssim (\sqrt{r} \|\hat G\| + \gamma) R \sqrt{\frac{\log(T(n+d))}{b}},\label{eq: T1 A4}
    \end{align}
    which holds with probability $1 - O((n+d)^{-1})$.
    For the term $T_2$, by triangle inequality and the inequality $\|A B\|_F \leq \|A\| \|B\|_F$ for any matrices $A$ and $B$, we obtain
    \begin{align}
        T_2^{(t)} &\leq \|(G_2^{(t)} - \hat G_2) \hat\Sigma_{x,\tilde x}^\top\|_F + \|(G_1^{(t)} - \hat G_1) \hat\Sigma_{x,\tilde x}\|_F + \alpha \|G^{(t)} G^{(t) \top} G^{(t)} - \hat G \hat G^\top \hat G\|_F + \alpha \|G^{(t)} - \hat G\|_F\nonumber\\
        &\leq \|G^{(t)} - \hat G\|_F \|\hat\Sigma_{x,\tilde x}\| + 3 \alpha (\|G^{(t)}\| \vee \|\hat G\|)^2 \|G^{(t)} - \hat G\|_F\nonumber\\
        &\lesssim \gamma \|\hat\Sigma_{x,\tilde x}\| + \alpha (\|\hat G\|^2 + \gamma^2) \gamma,\label{eq: T2 A4}
    \end{align}
    where we used $\|G^{(t)}\| \leq \|\hat G\| + \gamma$. The claim follows from \eqref{eq: T1 A4} and \eqref{eq: T2 A4}.
\end{proof}

\begin{proof}[Proof of Lemma \ref{lem: hessian}]
    The following proof uses the technique from the proof of Lemma B.7 in \citet{gao2021sparse}.
    We first derive the bound for the smoothness of $\mL$.
    To this aim, compute the derivatives of $\mL$. For notational brevity, we write $\Sigma$ for $\hat\Sigma_{x,\tilde x}$.
    Observe that
    \begin{align*}
        \pdv{}{G} \|G G^\top - I_r\|_F^2 = 4 G G^\top G - 4G.
    \end{align*}
    This implies
    \begin{align}
        \pdv{}{\vect(G)} \|G G^\top - I_r\|_F^2 = \vect(4 G G^\top G - 4G) = 4 (I_d \otimes G G^\top) \vect(G) - 4 \vect(G).\label{eq: first order derivative of penalty}
    \end{align}
    Write column vectors of $G$ as $G = [a_1, \dots, a_d] \in \R^{r \times d}$. Define
    \begin{align*}
        A \triangleq \begin{pmatrix}
            a_1 a^\top_1 & a_2 a^\top_1 & \dots & a_d a^\top_1\\
            a_1 a^\top_2 & a_2 a^\top_2 & \dots & a_d a^\top_2\\
            \vdots & \vdots & \ddots & \vdots\\
            a_1 a^\top_d & a_2 a^\top_d & \dots & a_d a^\top_d
        \end{pmatrix}.
    \end{align*}
    Thus,
    \begin{align}
        \pdv{\mL(G)}{G} &= - G \begin{pmatrix}
            O & \Sigma\\
            \Sigma^\top & O            
        \end{pmatrix} + \alpha G G^\top G - \alpha G.\label{eq: foc}
    \end{align}
    Also, \eqref{eq: first order derivative of penalty} further gives 
    \begin{align*}
        \frac{\partial^2}{\partial \vect(G) \partial \vect(G)^\top} \|G G^\top - I_r\|_F^2 = 4 (I_d \otimes G G^\top) + 4 G^\top G \otimes I_r + 4 A - 4 I_{rd}.
    \end{align*}
    Therefore,
    \begin{align*}
        \frac{\partial^2 \mL(G)}{\partial \vect(G) \partial \vect(G)^\top} &= - \begin{pmatrix}
            O & \Sigma\\
            \Sigma^\top & O            
        \end{pmatrix} \otimes I_r + \alpha I_{d} \otimes G G^\top + \alpha G^\top G \otimes I_r + \alpha A - \alpha I_{rd}.
    \end{align*}
    
    From Lemma \ref{lem: global minimum}, $\hat G = [\hat G_1, \hat G_2]$ is given by
    \begin{align*}
        \hat G_1 = \frac{1}{\sqrt{2}} V \qty(I_r + \frac{1}{\alpha} \Lambda_{[r]})^{1/2} P_{[r]}^\top, \ \ \hat G_2 = \frac{1}{\sqrt{2}} V \qty(I_r + \frac{1}{\alpha} \Lambda_{[r]})^{1/2} Q_{[r]}^\top,
    \end{align*}
    where $V \in \mathbb{O}_{r,r}$ is any orthogonal matrix, $P_{[r]}$ and $Q_{[r]}$ are the top-$r$ right and left singular vectors, respectively, and $\Lambda_{[r]}$ is a diagonal matrix of top-$r$ singular values.
    We note that $\|\hat G\|^2 \leq \|\hat G_1\|^2 + \|\hat G_2\|^2 = (1 + \|\Sigma\| / \alpha)$.
    
    Fix any $Z_1 \in \R^{r \times d_1}$ and $Z_2 \in \R^{r \times d_2}$.
    Let $Z = [Z_1, Z_2] \in \R^{r \times d}$. Write $z_1 \triangleq \vect(Z_1)$, $z_2 \triangleq \vect(Z_2)$ and $z = \vect(Z)$. Since
    \begin{align*}
        z^\top \qty(\begin{pmatrix}
            O & \Sigma\\
            \Sigma^\top & O            
        \end{pmatrix} \otimes I_r) z &= 2 z_1^\top (\Sigma \otimes I_r) z_2 = 2 \tr(Z_1^\top Z_2 \Sigma^\top),\\
        z^\top (I_d \otimes G G^\top) z &= \tr(Z^\top G G^\top Z),\\
        z^\top (G^\top G \otimes I_r) z &= \tr(Z^\top Z G^\top G),\\
        z^\top A z &= \sum_{i,j} a_j^\top z_i a_i^\top z_j = \tr(Z G^\top Z G^\top),
    \end{align*}
    we obtain
    \begin{align}
        z^\top \frac{\partial^2 \mL(G)}{\partial \vect(G) \partial \vect(G)^\top} z &= - 2 \tr(Z_1^\top Z_2 \Sigma^\top) - \alpha \tr(Z^\top Z)\nonumber\\
        &\quad+ \alpha \tr(Z G^\top Z G^\top) + \alpha \tr(Z^\top G G^\top Z) + \alpha \tr(Z^\top Z G^\top G).\label{eq: second order term}
    \end{align}
    Now suppose that $\|G - \hat G\|_F \leq \gamma$. By Cauchy-Schwarz inequality,
    \begin{align*}
        z^\top \frac{\partial^2 \mL(G)}{\partial \vect(G) \partial \vect(G)^\top} z &\leq 2 \|Z_1\|_F \|Z_2\|_F \|\Sigma\| + 3 \alpha \|Z\|_F^2 \|G\|^2\\
        &\leq (2 \|\Sigma\| + 3 \alpha \|G\|^2) \|Z\|_F^2.
    \end{align*}
    From $\|G\|^2 \leq (\|\hat G\| + \gamma)^2 \leq 2\|\hat G\|^2 + 2\gamma^2$, $\gamma^2 \leq 1$ and $\|\hat G\|^2 \leq (1 + \|\Sigma\|/\alpha)$, 
    \begin{align}
        z^\top \frac{\partial^2 \mL(G)}{\partial \vect(G) \partial \vect(G)^\top} z 
        &\leq \qty(8 \|\Sigma\| + 12 \alpha) \|Z\|_F^2.\label{eq: beta upper}
    \end{align}
    Setting $\beta_u \triangleq 8 \|\Sigma\| + 12 \alpha$ gives the first result for the smoothness of $\mL$.

    Next we derive the strong directional convexity of $\mL$.
    Let $\Delta_1 \triangleq H Z_1 - \hat G_1$, $\Delta_2 \triangleq H Z_2 - \hat G_2$ and $\Delta \triangleq [\Delta_1, \Delta_2] = H Z - \hat G$.
    We bound $\vect(\Delta)^\top \partial^2 \mL(G) \vect(\Delta)$ from below.
    We first deal with the case where $G = \hat G$.
    Since $\hat G \hat G^\top = (1/\alpha) V \Lambda_{[r]} V^\top + I_r$, \eqref{eq: second order term} gives,
    \begin{align*}
        &\vect(\Delta)^\top \frac{\partial^2 \mL(\hat G)}{\partial \vect(G) \partial \vect(G)^\top} \vect(\Delta)\\
        &\quad= - 2 \tr(\Delta_1^\top \Delta_2 \Sigma^\top) - \alpha \tr(\Delta^\top \Delta)\\
        &\quad\quad+ \alpha \tr(\Delta \hat G^\top \Delta \hat G^\top) + \alpha \tr(\Delta^\top \hat G \hat G^\top \Delta) + \alpha \tr(\Delta^\top \Delta \hat G^\top \hat G)\\
        &\quad= \underbrace{\tr(\Delta^\top V \Lambda_{[r]} V^\top \Delta) + \alpha \tr(\Delta^\top \Delta \hat G^\top \hat G) - 2\tr(\Delta_1^\top \Delta_2 \Sigma^\top)}_{=: T_1} + \underbrace{\alpha \tr(\Delta \hat G^\top \Delta \hat G^\top)}_{=: T_2}.
    \end{align*}
    We first bound the term $T_1$.
    Note
    \begin{align}
        \hat G^\top \hat G &= \frac{1}{2} \begin{pmatrix}
            P_{[r]}\\
            Q_{[r]}
        \end{pmatrix} \begin{pmatrix}
            P_{[r]}\\
            Q_{[r]}
        \end{pmatrix}^\top + \frac{1}{2\alpha} \begin{pmatrix}
            P_{[r]} \Lambda_{[r]} P_{[r]}^\top & P_{[r]} \Lambda_{[r]} Q_{[r]}^\top\\
            Q_{[r]} \Lambda_{[r]} P_{[r]}^\top & Q_{[r]} \Lambda_{[r]} Q_{[r]}^\top
        \end{pmatrix}.
    \end{align}
    Notice that the first term is positive semi-definite. Hence,
    \begin{align*}
        \alpha \tr(\Delta^\top \Delta \hat G^\top \hat G) &\geq \frac{1}{2} \tr(\Delta_1^\top \Delta_1 P_{[r]} \Lambda_{[r]} P_{[r]}^\top) + \tr(\Delta_1^\top \Delta_2 Q_{[r]} \Lambda_{[r]} P_{[r]}^\top) + \frac{1}{2} \tr(\Delta_2^\top \Delta_2 Q_{[r]} \Lambda_{[r]} Q_{[r]}^\top)\\
        &\geq 2 \tr(\Delta_1^\top \Delta_2 Q_{[r]} \Lambda_{[r]} P_{[r]}^\top),
    \end{align*}
    where we used $\tr(A B) \leq \|A\|_F \|B\|_F \leq (1/2) \tr(A^\top A) + (1/2) \tr(B^\top B)$, which follows from Cauchy-Schwarz inequality and $2xy \leq x^2 + y^2$.
    Let the SVD of $\Sigma$ be $\Sigma = P_{[r]} \Lambda_{[r]} Q_{[r]}^\top + P_\perp \Lambda_\perp Q_\perp^\top$,
    where $P_\perp$, $Q_\perp$ are the right and left singular vectors except top-$r$ singlar vectors, respectively, and $\Lambda_\perp$ is a diagonal matrix of remaining singular values.
    Observe that
    \begin{align*}
        &\alpha \tr(\Delta^\top \Delta \hat G^\top \hat G) - 2 \tr(\Delta_1^\top \Delta_2 \Sigma^\top)\\
        &\quad\geq 2 \tr(\Delta_1^\top \Delta_2 Q_{[r]} \Lambda_{[r]} P_{[r]}^\top) - 2 \tr(\Delta_1^\top \Delta_2 Q_{[r]} \Lambda_{[r]} P_{[r]}^\top) - 2 \tr(\Delta_1^\top \Delta_2 Q_\perp \Lambda_\perp P_\perp^\top)\\
        &\quad= - 2 \tr(\Delta_1^\top \Delta_2 Q_\perp \Lambda_\perp P_\perp^\top).
    \end{align*}
    Thus,
    \begin{align*}
        T_1 &\geq \tr(\Delta^\top V \Lambda_{[r]} V^\top \Delta) - 2 \tr(\Delta_1^\top \Delta_2 Q_\perp \Lambda_\perp P_\perp^\top)\\
        &\geq \lambda_r \|\Delta\|_F^2 - \tr(\Delta_2 Q_\perp \Lambda_\perp Q_\perp^\top \Delta_2^\top) - \tr(\Delta_1 P_\perp \Lambda_\perp P_\perp^\top \Delta_1^\top)\\
        &\geq (\lambda_r - \lambda_{r+1}) \|\Delta\|_F^2,
    \end{align*}
    where we used $\tr(A B) \leq (1/2) \tr(A^\top A) + (1/2) \tr(B^\top B)$ again.

    Next, we show $T_2 \geq 0$. 
    Suppose that $H Z \hat G^\top$ is symmetric. Then,
    \begin{align*}
        \tr(\Delta \hat G^\top \Delta \hat G^\top) &= \tr((H Z - \hat G) \hat G^\top (H Z - \hat G) \hat G^\top)\\
        &= \tr(H Z \hat G^\top H Z \hat G^\top) - 2 \tr(\hat G \hat G^\top H Z \hat G^\top) + \tr(\hat G \hat G^\top \hat G \hat G^\top)\\
        &= \tr((H Z \hat G^\top - \hat G \hat G^\top)^\top (H Z \hat G^\top - \hat G \hat G^\top)) \geq 0.
    \end{align*}
    Thus, we only need to show that $H Z \hat G^\top$ is symmetric.
    Recall that $H \triangleq \argmin_{O \in \mathbb{O}_{r,r}} \|O Z - \hat G\|_F^2$. This gives 
    \begin{align}
        \|H Z - \hat G\|_F^2 \leq \|H' Z - \hat G\|_F^2\label{eq: H1 inequality}    
    \end{align}
    for all $H' \in \mathbb{O}_{r,r}$. Let the SVD of $Z \hat G^\top$ be $Z \hat G^\top = U C V^\top$, where $U, V \in \mathbb{O}_{r,r'}$ are positive definite matrices and $C \in \R^{r'}$ is a diagonal matrix. 
    Write the orthogonal matrices of $U$ and $V$ as $U_\perp \in \mathbb{O}_{r,r-r'}$ and $V_\perp \in \mathbb{O}_{r,r-r'}$, respectively. From \eqref{eq: H1 inequality},
    \begin{align}
        \tr(V^\top H' U C) = \sum_{j \in [r']} (V^\top H' U)_{j,j} (C)_{j,j} \leq \tr(V^\top H U C) \ \ \ \ \forall H' \in \mathbb{O}_{r,r}.\label{eq: trace inequality}
    \end{align}
    The inequality in \eqref{eq: trace inequality} holds if and only if $V^\top H U = I_r$. Now we decompose $H U$ as $H U = VV^\top H U + V_\perp V_\perp^\top H U = V + V_\perp V_\perp^\top H U$. The fact that $V, H U \in \mathbb{O}_{r,r'}$ yields $H U = V$. Thus $H Z \hat G^\top = H U C V^\top = V C V^\top$ and hence $H Z \hat G^\top$ is symmetric.
    
    In summary, we showed that
    \begin{align}
        \vect(\Delta)^\top \frac{\partial^2 \mL(\hat G)}{\partial \vect(G) \partial \vect(G)^\top} \vect(\Delta) \geq (\lambda_r(\Sigma) - \lambda_{r+1}(\Sigma)) \|\Delta\|_F^2.\label{eq: hessian 1}
    \end{align}

    Next, we prove that $\vect(\Delta)^\top \partial^2 \mL(G) \vect(\Delta)$ is close to $\vect(\Delta)^\top \partial^2 \mL(\hat G) \vect(\Delta)$ under assumption $\|G - \hat G\|_F \leq \gamma$. Observe that
    \begin{align*}
        &\vect(\Delta)^\top \frac{\partial^2 \mL(G)}{\partial \vect(G) \partial \vect(G)^\top} \vect(\Delta) - \vect(\Delta)^\top \frac{\partial^2 \mL(\hat G)}{\partial \vect(G) \partial \vect(G)^\top} \vect(\Delta)\\
        &\quad= \alpha \tr(\Delta G^\top \Delta G^\top) + \alpha \tr(\Delta^\top G G^\top \Delta) + \alpha \tr(\Delta^\top \Delta G^\top G)\\
        &\quad\quad-\alpha \tr(\Delta \hat G^\top \Delta \hat G^\top) - \alpha \tr(\Delta^\top \hat G \hat G^\top \Delta) - \alpha \tr(\Delta^\top \Delta \hat G^\top \hat G).
    \end{align*}
    Using triangle inequality multiple times, we obtain
    \begin{align}
        &\abs{\vect(\Delta)^\top \frac{\partial^2 \mL(G)}{\partial \vect(G) \partial \vect(G)^\top} \vect(\Delta) - \vect(\Delta)^\top \frac{\partial^2 \mL(\hat G)}{\partial \vect(G) \partial \vect(G)^\top} \vect(\Delta)}\nonumber\\
        &\quad\leq 3 \alpha \|\Delta\|_F^2 \|G - \hat G\|_F (\|G\| + \|\hat G\|)\nonumber\\
        &\quad\leq 3 \alpha \|\Delta\|_F^2 \|G - \hat G\|_F (\|G - \hat G\| + 2\|\hat G\|)\nonumber\\
        &\quad\leq 3 \alpha \|\Delta\|_F^2 \gamma \qty(1 + 2 \qty(1 + \frac{\|\Sigma\|}{\alpha})^{1/2})\nonumber\\
        &\quad\leq 9 \alpha \|\Delta\|_F^2 \gamma \qty(1 + \frac{\|\Sigma\|}{\alpha})^{1/2}\nonumber\\
        &\quad\leq \frac{\lambda_r(\Sigma) - \lambda_{r+1}(\Sigma)}{2} \|\Delta\|_F^2,\label{eq: hessian 2}
    \end{align}
    where the second last inequality follows from the assumption and $\|G - \hat G\| \leq \|G - \hat G\|_F \leq \gamma \leq 1$,
    and the last inequality follows from \eqref{eq: choice of gamma}.
    Combining \eqref{eq: hessian 1} and \eqref{eq: hessian 2} gives
    \begin{align*}
        \vect(\Delta)^\top \frac{\partial^2 \mL(G)}{\partial \vect(G) \partial \vect(G)^\top} \vect(\Delta) &\geq \frac{\lambda_r(\Sigma) - \lambda_{r+1}(\Sigma)}{2} \|\Delta\|_F^2.
    \end{align*}
    Setting $\beta_l \triangleq (\lambda_r(\Sigma) - \lambda_{r+1}(\Sigma)) / 2$ concludes the proof.
\end{proof}

\begin{proof}[Proof of Lemma \ref{lem: global minimum}]
    Here we derive the minimizer of $\mL'$.
    Let the singular value decomposition of $\Sigma$ be $\Sigma = P \Lambda Q^\top$, where $\Lambda \in \R^{d_1 \times d_2}$ is a diagonal matrix and $P = (p_1, \dots, p_{d_1}) \in \mathbb{O}_{d_1,d_1}$ and $Q = (q_1, \dots, q_{d_2}) \in \mathbb{O}_{d_2,d_2}$ are orthogonal matrices.
    By setting \eqref{eq: foc} to be $0$, 
    we obtain
    \begin{align*}
        G \begin{pmatrix}
            O & \Sigma\\
            \Sigma^\top & O
        \end{pmatrix} = \alpha (GG^\top - I_r) G.
    \end{align*}
    Equivalently,
    \begin{align}
        G_1 \Sigma &= \alpha (G G^\top - I_r) G_2,\label{eq: G1 Sigma}\\
        G_2 \Sigma^\top &= \alpha (G G^\top - I_r) G_1.\label{eq: G2 Sigma}
    \end{align}
    Multiplying $\Sigma$ from the right in \eqref{eq: G2 Sigma}, and substituting with \eqref{eq: G1 Sigma} gives
    \begin{align*}
        G_2 \Sigma^\top \Sigma = \alpha (G G^\top - I_r) G_1 \Sigma = \alpha (G G^\top - I_r)^2 G_2.
    \end{align*}
    Thus the right singular vectors of $G_2$ is aligned with some $r$ column vectors of $Q$. 
    Given an indices set $J = \{j_1, \dots, j_r\}$, we write $\Lambda_{J} \triangleq \diag((\Lambda)_{j_1, j_1}, \dots, (\Lambda)_{j_r, j_r})$, $P_J \triangleq (p_{j_1}, \dots, p_{j_r})$ and $Q_J \triangleq (q_{j_1}, \dots, q_{j_r})$.
    We decompose $G_2$ by SVD as $G_2 = V_2 C_2 U_2^\top$, where $U_2 = Q_I$ for some $I = (i_1, \dots, i_r) \subset [d_1]$ and $C_2 \in \R^{r \times r}$ is a diagonal matrix, and $V_2 \in \mathbb{O}_{r,r}$.
    Similarly, we decompose $G_1$ as $G_1 = V_1 C_1 U_1^\top$, where $U_1 = P_{I'}$ for some $I' = (i_1', \dots, i_r') \subset [d_2]$, $C_1 \in \R^{r \times r}$ is a diagonal matrix, and $V_1 \in \mathbb{O}_{r,r}$.
    
    From \eqref{eq: G1 Sigma}, we obtain
    \begin{align*}
        V_1 C_1 \Lambda_{I'} Q_{I'}^\top = \alpha (G G^\top - I_r) V_2 C_2 Q_I^\top.
    \end{align*}
    Thus $Q_{I'} = Q_I H_Q$ for some $H_Q \in \mathbb{O}_{r,r}$. Without loss of generality, we set $I = I'$.
    
    Since the term $-\tr(G_1 \hat\Sigma G_2^\top) = -\tr(V_1 C_1 \Lambda_I C_2 V_2^\top)$ in the loss function is minimized when $V_1 = V_2$, whereas the penalty term $\Pi(G_1, G_2)$ is invariant under the change of $V_1$ and $V_2$, we obtain $V_1 = V_2$.
    In summary, from \eqref{eq: G1 Sigma} and \eqref{eq: G2 Sigma}, we have
    \begin{align*}
        V_1 C_1 \Lambda_I Q_I^\top &= \alpha V_1 (C_1^2 + C_2^2 - I_r) C_2 Q_I^\top,\\
        V_1 C_2 \Lambda_I P_I^\top &= \alpha V_1 (C_1^2 + C_2^2 - I_r) C_1 P_I^\top.
    \end{align*}
    Thus
    \begin{align}
        C_1 \Lambda_I &= \alpha (C_1^2 + C_2^2 - I_r) C_2, \ \ C_2 \Lambda_I = \alpha (C_1^2 + C_2^2 - I_r) C_1.\label{eq: C1 and C2}
    \end{align}
    Fix any $j \in [r]$. Suppose that $j$-th entry of $C_1$ is $0$. Then, from \eqref{eq: C1 and C2}, $j$-th entry of $C_2$ must be $0$. Now the loss function can be written as
    \begin{align*}
        \mL = -\tr(C_1 \Lambda_I C_2) + \frac{\alpha}{4} \|C_1^2 + C_2^2 - I_r\|_F^2.
    \end{align*}
    Note that we can make the loss smaller by slightly increasing $j$-th diagonal entry of $C_1$ and $C_2$. This implies that $j$-th diagonal entry of $C_1$ and $C_2$ cannot be $0$ when $G_1$ and $G_2$ are the solution to the minimization of the loss $\mL(G)$. 
    Since $j$ is arbitrary, we can show that $C_1 = C_2 = (1/\sqrt{2}) (I_r + (1/\alpha) \Lambda_I)^{1/2}$.
    
    Next we show $I = [r]$. To see this, note that
    \begin{align*}
        \mL &= -\frac{1}{2} \tr( \qty(I_r + \frac{1}{\alpha} \Lambda_I)^{1/2} \Lambda_I \qty(I_r + \frac{1}{\alpha} \Lambda_I)^{1/2} ) + \frac{\alpha}{4} \tr((C_1^2 + C_2^2 - I_r)^2)\\
        &= - \frac{1}{2}\sum_{i \in I} \qty(\lambda_i + \frac{\lambda_i^2}{\alpha}) + \frac{1}{4 \alpha} \sum_{i \in I} \lambda_i^2\\
        &= - \frac{1}{2} \sum_{i \in I} \qty(\lambda_i + \frac{\lambda_i^2}{2\alpha}),
    \end{align*}
    which is minimized if and only if $I = [r]$ due to the assumption $\lambda_r(\Sigma) > \lambda_{r+1}(\Sigma)$.
    Finally, $\hat G$ is given by
    \begin{align}
        \hat G_1 = \frac{1}{\sqrt{2}} V \qty(I_r + \frac{1}{\alpha} \Lambda_{[r]})^{1/2} P_{[r]}^\top, \ \ \hat G_2 = \frac{1}{\sqrt{2}} V \qty(I_r + \frac{1}{\alpha} \Lambda_{[r]})^{1/2} Q_{[r]}^\top,\label{eq: optimal G}
    \end{align}
    where $V \in \mathbb{O}_{r,r}$ is any orthogonal matrix.
\end{proof}

\section{Auxiliary Results}

Here we define Orlicz norm of a random variable $X$ as $\|X\|_{\psi_2} \triangleq \inf \{c > 0: \mathbb{E}[e^{X^2/c^2}] \leq 2\}$.

\begin{assumption}\label{asm: cross-covariance concentration ap}
    Let $X$ and $\tilde X$ be mean zero random vectors taking values in $\R^{d_1}$ and $\R^{d_2}$, respectively.
    Assume that there exists some constants $C_1, C_2 > 0$ satisfying that $\mathbb{E}[(u^\top X)^2] \geq C_1 \|u^\top X\|_{\psi_2}^2$ holds for any $u \in \R^{d_1}$, and that $\mathbb{E}[(v^\top \tilde X)^2] \geq C_2 \|v^\top \tilde X\|_{\psi_2}^2$ holds for any $v \in \R^{d_2}$.
\end{assumption}

We borrow the proposition from \citet{nakada2023understanding} for bounding the distance between sample cross-covariance matrix and population cross-covariance matrix.
\begin{lemma}[Proposition 9.1 from \citet{nakada2023understanding}]\label{lem: cross-covariance concentration}
    Suppose that Assumption \ref{asm: cross-covariance concentration ap} holds. 
    Let $(X_1, \tilde X_1), \dots, (X_n, \tilde X_n)$ be independent copies of $(X, \tilde X)$.
    Let $\hat\Sigma_{X,\tilde X} \triangleq (1/n) \sum_{i=1}^n X_i \tilde X_i^\top$.
    Then, there exists some constant $C = C(C_1, C_2) > 0$ such that with probability at least $1 - e^{-t}$,
    \begin{align*}
        &\|\hat\Sigma_{X,\tilde X} - \mathbb{E}[\hat\Sigma_{X,\tilde X}]\|\\
        &\quad\leq C \qty[ (\tr(\Sigma_{\tilde X}) \|\Sigma_X\| \vee \tr(\Sigma_X) \|\Sigma_{\tilde X}\|)^{1/2} \sqrt{\frac{t + \log(d_1 + d_2)}{n}} \vee (\tr(\Sigma_X) \tr(\Sigma_{\tilde X}))^{1/2} \frac{t + \log(d_1 + d_2)}{n} ]
    \end{align*}
    holds for all $t > 0$.
\end{lemma}

\begin{lemma}[Corollary 2.5 from \citet{bardenet2015concentration}]\label{lem: subsampling concentration}
    Let $X_1, \dots, X_n \in \R$ be fixed numbers.
    Let $\mB \subset [n]$ be a random batch of size $b$ from $[n]$ without replacement. Then,
    \begin{align*}
        \abs{\frac{1}{b} \sum_{i \in \mB} X_i - \frac{1}{n} \sum_{i \in [n]} X_i} \lesssim \max_{i \in [n]} |X_i| \sqrt{\frac{(1 - b/n) \log(1/\gamma)}{b}}
    \end{align*}
    holds with probability $1 - \gamma$.
\end{lemma}

\begin{lemma}[Modification of Lemma C.1 from \citet{nakada2023understanding}]\label{lem: good event}
    Suppose Assumptions \ref{asm: signal condition number ap} and \ref{asm: snr ap} hold.
    Then, the following inequalities hold with probability $1 - O((n+d)^{-1})$:
    \begin{align*}
        &\max_{i \in [n]} \|x_i\| \leq C_1 \|\Sigma_z\|^{1/2} \sqrt{(r + s_1^{-2} r_e(\Sigma_\xi)) \log (n+d)},\\
        &\max_{i \in [n]} \|\tilde x_i\| \leq C_2 \|\Sigma_{\tilde z}\|^{1/2} \sqrt{(r + s_2^{-2} r_e(\Sigma_{\tilde \xi})) \log (n+d)},
    \end{align*}
    where $C_1' = C_1'(\sigma, s_1), C_2' = C_2'(\sigma, s_2)$ are some constants.
\end{lemma}

\end{document}